\documentclass[10pt,twoside,twocolumn]{IEEEtran}

\ifCLASSINFOpdf
   \usepackage[pdftex]{graphicx}
   \DeclareGraphicsExtensions{.pdf,.jpg,.png}
\else
\fi

\usepackage[scriptsize,textfont=sf]{caption}

\usepackage[cmex10]{amsmath}
\usepackage{array}

\usepackage{mdwmath}
\usepackage{mdwtab}
\usepackage{eqparbox}

\usepackage{xspace}
\usepackage{times}
\usepackage{graphicx}
\usepackage{amsmath}
\usepackage{amssymb}
\usepackage{yhmath}
\usepackage{subfigure}
\usepackage{slashbox,pict2e}
\usepackage{xcolor}

\usepackage{algorithm2e}
\let\savedalgorithm\algorithm
\let\savedendalgorithm\endalgorithm
\newenvironment{algorithmic}{%
\savedalgorithm
}{%
\savedendalgorithm
}

\usepackage{algorithm}
\makeatletter
\DeclareRobustCommand\onedot{\futurelet\@let@token\@onedot}
\def\@onedot{\ifx\@let@token.\else.\null\fi\xspace}

\def\eg{\emph{e.g}\onedot} 
\def\ie{\emph{i.e}\onedot}

\def\etal{\emph{et al}\onedot}
\makeatother

\def\mm{{{$\boldsymbol m$}$-$}\xspace}

\def\cII{{double-minima}\xspace} 
\def\cI{{single-minimum}\xspace} \def\CI{{single-minimum}\xspace}

\def\rl{{{\widetilde l}}}

\begin{document}

\title{A Computational Model of the Short-Cut Rule for 2D Shape Decomposition}

\author{Lei~Luo,~Chunhua~Shen,~Xinwang~Liu,~Chunyuan~Zhang
\thanks{L.~Luo, X.~Liu and C.~Zhang are
with the College of Computer, National University of Defense Technology,
Changsha, Hunan, 410073, China.
 (e-mail: \{l.luo, cyzhang\}@nudt.edu.cn).
 Part of the work was done when L. Luo was visiting The University of Adelaide.
}
 \thanks{C.~Shen is with
   Australian Research Council Centre of Excellence for Robotic Vision,
   The University of Adelaide, Adelaide, SA 5005, Australia
 (e-mail: chunhua.shen@adelaide.edu.au).}
}

\markboth{Manuscript}%
{L.~Luo \MakeLowercase{\textit{et al.}} }

\maketitle

\begin{abstract}

    We propose a new 2D shape decomposition method based on the short-cut rule.
    The short-cut rule originates from cognition research, and states that the
    human visual system prefers to partition an object into parts using the
    shortest possible cuts. We propose and implement a computational model for
    the short-cut rule and apply it to the problem of shape decomposition.  The
    model we proposed generates a set of cut hypotheses passing through the
    points on the silhouette which represent the negative minima of curvature.
    We then show that most part-cut hypotheses can be eliminated by analysis of
    local properties of each. Finally, the remaining hypotheses are evaluated
    in ascending length order, which guarantees that of any pair of conflicting
    cuts only the shortest will be accepted. We demonstrate that, compared with
    state-of-the-art shape decomposition methods, the proposed approach
    achieves decomposition results which better correspond to human intuition
    as revealed in psychological experiments.

\end{abstract}


\section{Introduction}

    \IEEEPARstart{C}{ognition} research suggests that
    the human visual system represents
    the shape of an object in terms of a set of parts and the spatial
    relationships between them \cite{Hoffman2,Hoffman1}.
    While parts-based approaches have been very successful in
    problems such as object recognition,
    shape simplification, skeleton extracting and collision detection, the problem of
    decomposing a shape into a suitable set of parts remains  challenging.

  In psychophysics study, several hypotheses, \eg, \emph{the minima rule}
  \cite{m-8} and \emph{the short-cut rule} \cite{shortcut4}, have been
  suggested to understand the shape decomposition process of human visual
  system. The minima rule points out that human vision defines part boundaries
  at points of \emph{negative minima of curvature} (\mm for short as in
  \cite{Psych9}) on the silhouette. It indicates the strong link between
  visually meaningful parts and near-convex geometries. As stated by Basri
  \etal \cite{similarity11}, parts generally are defined to be convex or nearly
  convex shapes separated from the rest of the object at concavity extreme.
  Therefore, a range of approaches try to decompose the shape into near-convex
  parts \cite{DCE7, convex12, CVPR6, 2011ICCV_NearConvex}. Besides visual
  parts, skeletons are also conjectured to be the intermediate-level
  representation of objects \cite{nature13}.
  Thus, another category of approaches combine the previous hypotheses and
  skeletons in shape decomposition \cite{2008BoneGraph, ICCV5, pami3}.
  {\em However,
  the short-cut rule, which specifies that the human visual system prefers to
  connect segmentation points that are in close proximity to form a part, is
  less considered, or even absent, in both categories.
  }

  Here we propose a shape decomposition method based on the psychophysics
  studies, especially the short-cut rule. The method can be seen as a
  computational model for the short-cut rule, and thus represents an attempt to
  devise a shape partitioning system grounded in psychophysics research. The
  method represents the first practicable algorithm for implementing the
  short-cut rule, and the first application of the short-cut rule to practical
  shape decomposition.
The short-cut rule states that the human visual system prefers to partition
a shape into parts using shorter cuts when
other conditions are equal. In other words, if there are two
conflicting candidate part cuts that might be applied to a shape,
humans are more likely to ``see'' the shorter cut and hence
reject the longer alternative.
  The proposed algorithm consists of two steps:
\begin{enumerate}
  \item First,
        we implement the minima-rule by constructing part-cut hypotheses from
        concave vertexes of the simplified shape polygon---an approximation of
        the shape contour. Following the idea of \emph{ligature} and
        \emph{semi-ligature} in \cite{1999Ligature} and \cite{1978Blum}, we
        propose to partition the set of cuts into two classes according to the
        number of \mm endpoints that they have, where a \emph{double-minima
        cut} has two \mm endpoints and a \emph{single-minimum cut} has only one
        \mm endpoint. Then, we derive a set of heuristic constraints from
        psychophysics rules and visual observations to largely reduce the
        number of the two classes of part-cut hypotheses.
  \item Second, we determine part-cuts in a greedy fashion based on the
short-cut rule that examines part-cut hypotheses in ascending order of their
\emph{relative lengths}.
\end{enumerate}
As we discuss below, in this setting,
there are at most two part-cuts for every \mm point. As a result of this
observation, it is typically possible to terminate the process  before all
candidates have to be examined, which also simplifies the process of conflict
resolution between part-cut hypotheses.

\begin{figure}[t]
    \begin{center}
    \subfigure[]{\includegraphics[width=.115\textwidth]{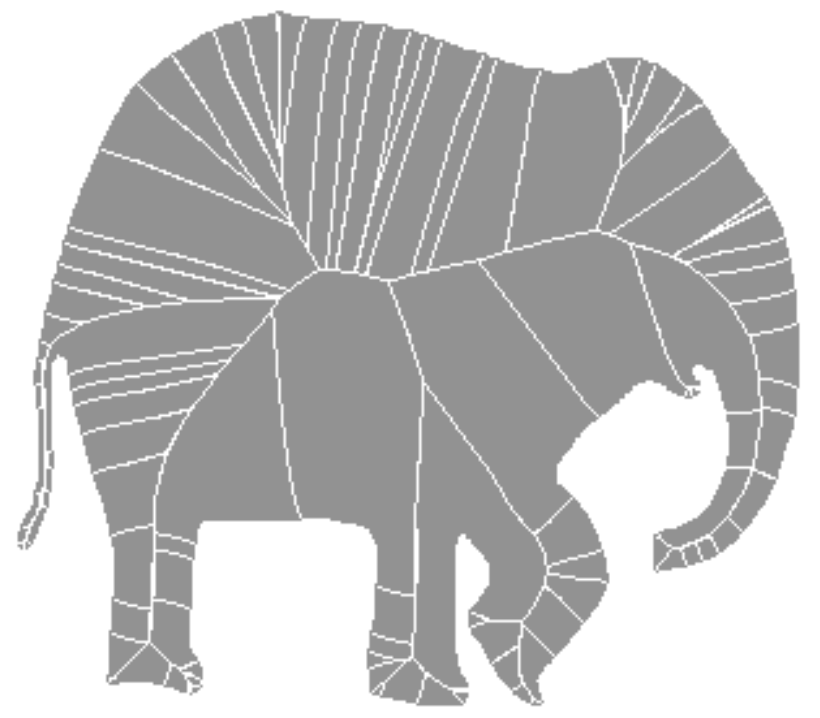}}
    \subfigure[]{\includegraphics[width=.115\textwidth]{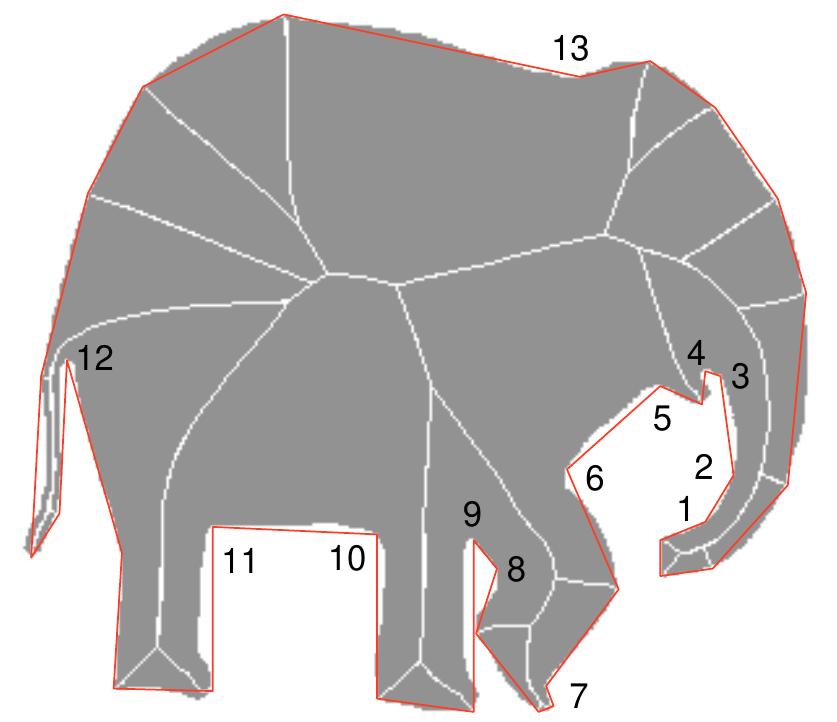}}
    \subfigure[]{\includegraphics[width=.115\textwidth]{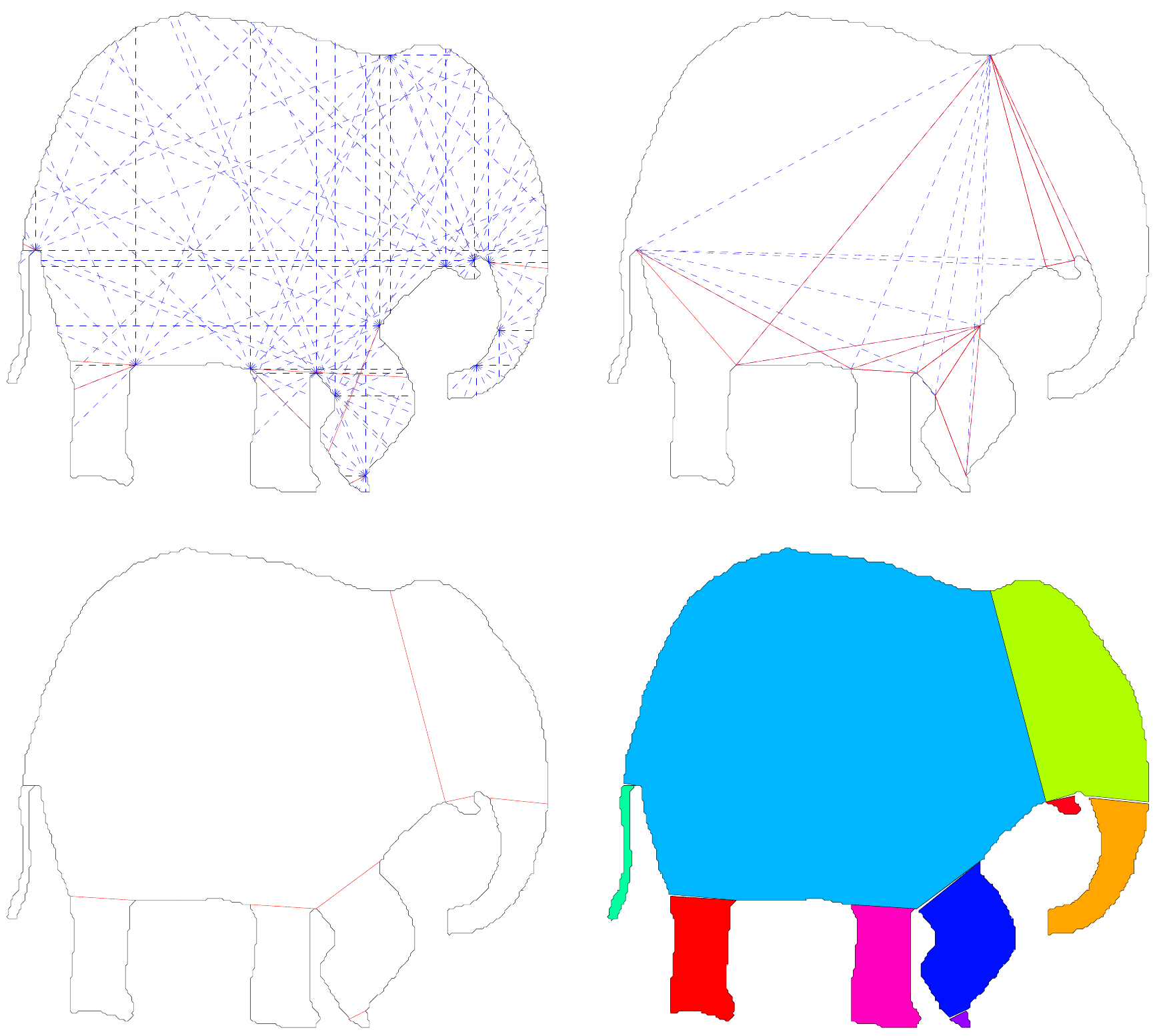}}
    \subfigure[]{\includegraphics[width=.115\textwidth]{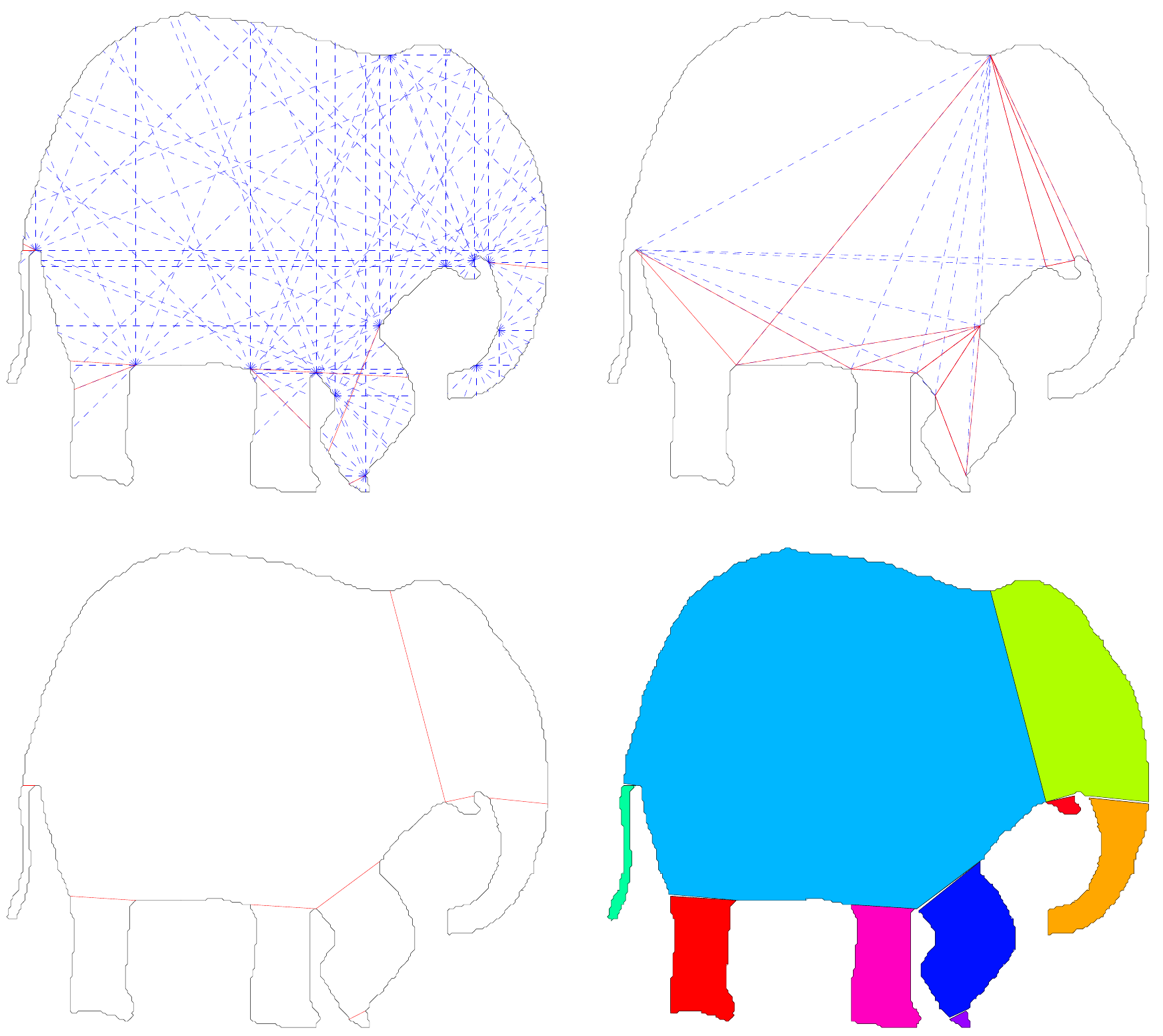}}\\
    \vspace{-.2cm}
    \subfigure[]{\includegraphics[width=.115\textwidth]{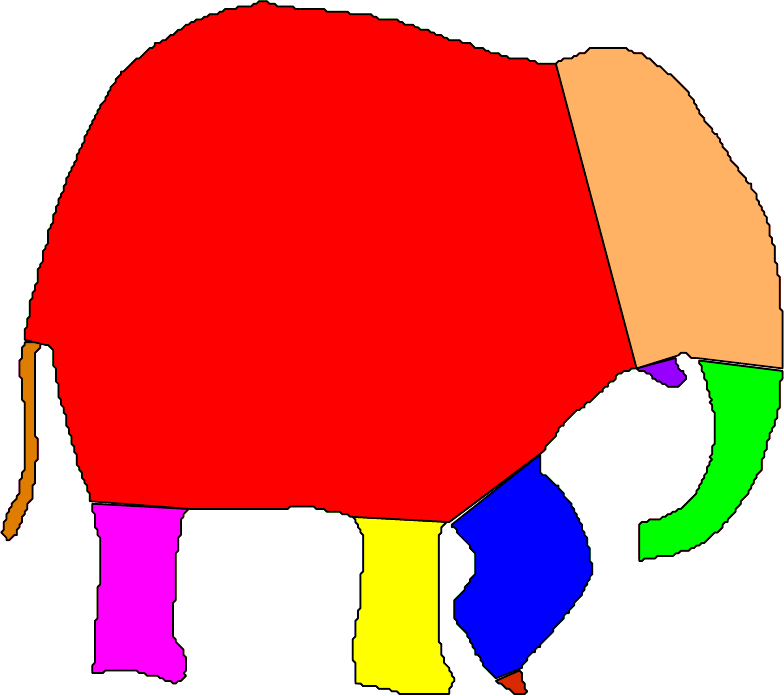}}
    \subfigure[]{\includegraphics[width=.115\textwidth]{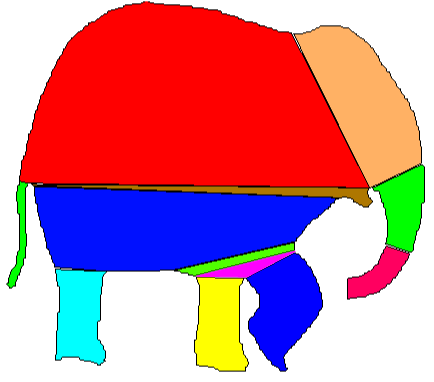}}
    \subfigure[]{\includegraphics[width=.115\textwidth]{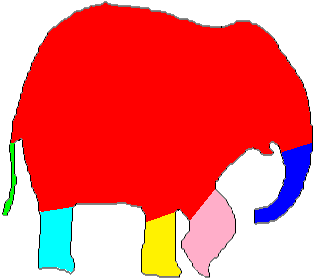}}
    \subfigure[]{\includegraphics[width=.115\textwidth]{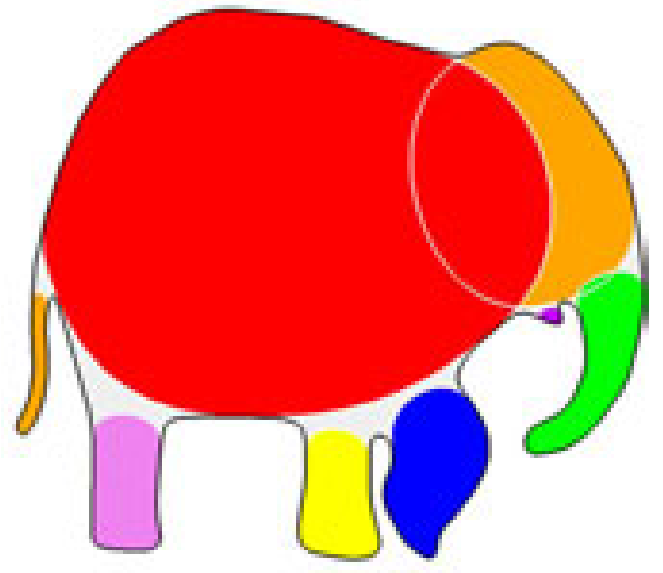}}
    \end{center}
    \vspace{-.4cm}
    \caption{
       The decomposition process of an elephant shape (best viewed on screen).
       (a) The silhouette and skeleton of an elephant.
       (b) The simplified shape polygon (in red solid line) after using DCE,
       where the 13 concave vertexes are served as \mm points of the origin shape contour.
       This process also helps in pruning redundant skeleton branches \cite{bai2007skeleton}.
       (c)-(d) \CI and \cII part-cut hypotheses, those discarded and remaining ones are shown as blue dashed and red solid lines, respectively.
       (e)-(f) Decomposition results of the proposed method, \cite{convex12}, \cite{CVPR6} and \cite{ICCV5}, respectively.
    }
    \label{fig:elephant_whole}
    \end{figure}

  An illustration of the proposed shape decomposition process is shown in Fig.
  \ref{fig:elephant_whole}. Given the target shape, an elephant, represented by
  a silhouette in (a), the outline of it is  a closed polygon and is
  simplified to have fewer vertexes by using Discrete Curve Evolution (DCE, see
  Appendix A for a brief description). As shown in Fig.
  \ref{fig:elephant_whole}(b), the DCE process removes those redundant skeleton
  branches generated by noise such that the stable skeletons and the corresponding
  perceptual visual parts present more clearly. Then, the set of \cI
  part-cut hypotheses in (c) and the set of \cII part-cut hypotheses in (d) are
  separately constructed from those concave vertexes ((in this example, 13 vertexes)
    that represent the
  negative minima of curvature.
  Finally, the part-cuts are selected from the hypotheses sets by a greedy
  algorithm, and decompose the shape into a few (in this case, 9) visual parts in (e).

  The proposed method builds upon solid psychophysics principles, and experimental results show
  that our
  decomposition results can better
      better correspond to human intuition as revealed in psychological experiments,
  comparing with
  those methods that only decompose shapes into near-convex parts.
  See our result and other results on the ``elephant'' shape
  in (e)-(f) in Fig. \ref{fig:elephant_whole}.

  The other category of decomposition methods such as \cite{ICCV5}
  may have taken psychophysics principles into consideration too.
  However, the work of \cite{ICCV5} depends on the skeleton or symmetric axes of
  the shape, which are typically computationally expensive to obtain.
  Our proposed approach avoids
  that using approximation methods based on the local geometry of a part-cut
  hypothesis.

    We make two major contributions in this work.
    \begin{enumerate}
      \item First, we devise a computational procedure for the short-cut rule
        from psychophysics. Using this procedure, we propose a new approach to
        2D shape decomposition. Compared with previous shape decomposition
      methods \cite{convex12,CVPR6,ICCV5}, our method obtains more intuitive
    decomposition results.
  \item The second contribution is that we
      discriminate all cuts into two types following the work of
      \cite{1999Ligature}, which helps in generating part-cut hypotheses and
      discarding meaningless cuts.
  \end{enumerate}

    Furthermore, we define a quantitative evaluation for shape decomposition
    method based on the psychological study of \cite{Psych9}. Previously the
    quality of shape decomposition is often subjectively judged by inspection
    of a small number of decomposition results.

    Next, we review some work that is most relevant to  ours and then present
    our main results. We show the experiments in Section~\ref{sec:exp} and
    conclude the paper in Section~\ref{sec:con}.

\subsection{Related work}

	The majority of existing shape decomposition approaches can be classified into two categories.
	One category aims to decompose shapes into near-convex parts.
	The other category tries to decompose shapes into natural parts based on psychophysics studies.

	The first category usually decomposes shapes based on the convexity constraint.
	This is mainly because that convexity plays an important role in human perception \cite{similarity11}, and is also supported by the minima rule.
	Conventional strict convex decomposition is a well-studied problem, but is not suitable in the case of shape decomposition.
	Because human may ``see" a shape in different scales to find the most perceivable meaning of it.
	When we decompose a shape at a particular scale, non-convexity below that scale should be neglected.
	Latecki and Lakamper \cite{DCE7} thus developed the DCE algorithm to control
  the tolerance level of non-convexity,
  and decompose shapes with concave vertexes of the DCE-simplified shape polygon.
	Lien \etal \cite{convex12,lu2012alpha,ghosh2013fast} proposed Approximate Convex Decomposition (ACD),
  which decomposes shapes into approximately convex parts.
	Liu \etal \cite{CVPR6} proposed Convex Shape Decomposition (CSD) by formulating
  the shape decomposition problem as an integer linear programming problem,
	which is further extended in \cite{2011ICCV_NearConvex, jiang2013toward, ICCV2013Ma},
  by  introducing visual naturalness regularization terms into the object function.

	The other category aims to decompose shapes into natural parts.
	Skeletons or local symmetric axes of shapes are usually involved according to the psychological model in \cite{nature13}.
	Singh and Hoffman \cite{Hoffman1} defined a part-cut,
  the border line of two parts, as a straight line inside the shape that crosses an axis of local symmetry.
	Macrini \etal \cite{2008BoneGraph,1999Ligature} recognized the ligature in Blum's skeleton \cite{1978Blum}
  as the glue between parts and represented the part structure of a shape by a bone graph.
    Probabilistic approaches such as \cite{zhu1999stochastic,feldman2006bayesian}
    were proposed to provide a more robust estimation of the skeletal structure,
    which more faithfully represent the part structure.
    Other types of skeletons,
    such as \emph{shocks} \cite{shock14} and \emph{smoothed local
    symmetries} (SLS) \cite{SLS15}, have also been used.
	The \emph{limb} and \emph{neck} based method of \cite{pami3} used shocks to
    extract necks and boundary curvature to find limbs.
    Mi and DeCarlo \cite{ICCV5} argued that shapes should be decomposed by
    \emph{part transitions} instead of part-cuts. They traced along
    the axes of SLS to find regions with strong transitional strength
    and decomposed shapes into overlapped parts.
    We have borrowed the idea of part transitions to define the \emph{neighborhood histogram},
    which is essential to our decomposition approach.

\section{Our Approach}
\label{sec:main}

    The psychological study in \cite{Psych9} suggests that most
    subjects segment an object shape into parts on the basis of
    straight lines that start and end on the boundary, which are
    called the ``part-cuts" of the shape.  According to the conditions
    laid out in \cite{Hoffman1}, a part-cut is ``a straight line
    segment'' that joins two points on the outline of a silhouette such
    that the following conditions are satisfied:
    \begin{itemize}
        \item[\upshape(\itshape a\upshape)] at least one of the two
    points has negative curvature;
        \item[\upshape(\itshape b\upshape)]
    the entire segment locates in the interior
    of the shape; and
        \item[\upshape(\itshape c\upshape)]
    the segment crosses an axis of local symmetry.
    \end{itemize}
    Given a shape $X \subset \mathbb{R}^2 $, together with a set of part-cuts
    $C = \{ c_1 ,c_2 ,...,c_N \} $.  The shape can be written as
    $X = \{ P_1,P_2,...,P_{N+1} \}$, where $\forall i,1 \leqslant i \leqslant N + 1,P_i  \subset \mathbb{R}^2 $
    is a visual part of $X$, and $P_i  \cap P_j  = c_{ij}  \in C
    {\text{ or }} P_i  \cap P_j=\emptyset,\forall i \ne j,1 \leqslant
    i,j \leqslant N + 1$. The problem that we are interested is how to obtain part-cuts $C$.
    The definition of ``part-cuts" in \cite{Hoffman1} thus guides our work.
    We firstly find segmentation points on the basis of
    Condition ($a$), and then extract a set of part-cut hypotheses
    $H=\{h_1,h_2,...,h_{|H|}\}$  that are consistent with Condition ($b$).
    Here $ |H|$ is the size of the set $ H $.
    The set of part-cuts $C$ is chosen from $H$
    with minimal total cost  based on Condition ($c$) and the short-cut rule:
    \begin{equation}\label{formula:f1}
        \mathop {\min }\limits_{ C} \sum\limits_{i = 1}^N {{\text{Cost}}(c_i )},
    \end{equation}
    where $c_i \in C$, and Cost($c_i$) is a positive quantity that represents
    the cost associated with the part-cut $c_i$.

    \SetKwInput{KwInit}{Initialization}
    \SetVline
    \linesnumbered
    \begin{algorithm}[t]
    \caption{Shape decomposition.}
    \begin{algorithmic}
    	\KwIn{Shape $X$;}
    	\KwOut{Part-cuts $C$;}
    	Extract part-cut hypotheses by Algorithm 2;\\
    	Obtain part-cuts $C$ by Algorithm 3;
    \end{algorithmic}
    \label{alg:a1}
    \end{algorithm}

    The overall framework of our algorithm
    is shown in Algorithm~1.
    Finally, the shape $X$  is decomposed into visual parts by part-cuts in $C$.

\subsection{Finding part-cut hypotheses}

    In this section, we present our approach for finding the part-cut hypotheses.

\subsubsection{Segmentation points}
\label{sec:211}

    The minima rule suggests that human vision separates shape contours at
    \mm points. Condition ($ a $) above
	relaxes this constraint by allowing
    other geometric factors to play a role in parsing.  Winter and Wagemans'
    large-scale experiments \cite{Psych9} revealed that  nearly $70\%$ of segmentation points
    are \mm points.
	Based on
    this fact, {\em we use \mm as segmentation points}. Since the contours of
    shapes are usually distorted by noise, we simplify the contour with
    a closed polygon in order to find \mm points.
    This process does not significantly affect the final
    decomposition result as visual perception inherently allows some degree
    of deviation. DCE of \cite{DCE7} is one of the contour
    simplification methods that
    preserves perceptual appearance while eliminating distortions.
    As shown in Fig.\ \ref{fig:elephant_whole},
    the noise in the outline of the top left silhouette generates
    a lot of redundant skeleton branches.
{
     After applying DCE, the contour of the elephant is represented as a polygon, in which
     the redundant skeleton branches are removed.
}
    As we will discuss later, the proposed algorithm is closely related to the skeleton
    (local symmetric axis) although
	the skeleton itself is never explicitly calculated.
	Within the polygonal silhouette generated by
	DCE, every concave vertex corresponds to a curvature minimum of the original shape.
    Rather than consider every such vertex, however,
    we introduce a threshold $\delta$, which is the minimal degree of concavity
    such that a vertex $p$ is considered to be an \mm point only if its interior angle
    $\angle p > \pi + \delta$.
    This strategy simplifies computation and more importantly,
    tolerates the approximation error introduced by DCE.
    Taking perceptual salience and approximation error into account,
    we set $\delta= \frac{ \pi } {  9 }$ in all of our experiments.

\subsubsection{Two types of cuts}

    \begin{figure}[t]
    \begin{center}
    \includegraphics[width=0.48\textwidth]{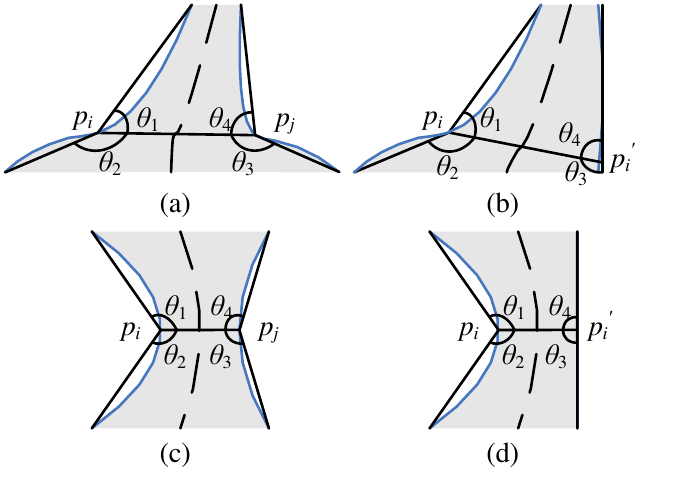} %
    \end{center}
    \vspace{-.4cm}
       \caption{
       Angles formed by (a) limb, (b) neck,
       (c) limb-like \cI cuts, (d) neck-like \cI cuts and the simplified shape polygons.
       Limb and neck are both \cII cuts.
       The left parts of the two shapes in (a), (b) and (c), (d) share the same geometry.
       $p_ip_j$ and $p_ip_i'$ are of the same length.
       }
    \label{fig:crossAxis}
    \end{figure}

    August and Siddiqi in \cite{1999Ligature} suggested that the \emph{ligature} and
    \emph{semi-ligature} of \cite{1978Blum} serve as the ``glue" between parts,
    where a ligature related to two \mm points defines a limb or a neck
    and a semi-ligature related to a single \mm point divides a tail
    from the main body of the shape
    (see the elephant's tail in Fig.~\ref{fig:elephant_whole} as an example).
	On this basis we partition the set of cuts into two classes
    based on the number of \mm endpoints that they have.
    As shown in Fig.~\ref{fig:crossAxis}, limbs and necks, which have two \mm endpoints, are called \emph{\cII cuts}.
    In contrast a \emph{\cI cut} has only one \mm endpoint.
    Following the idea of limb and neck, we further divide \cI cuts into neck-like \cI cuts and limb-like \cI cuts.

    The two endpoints of a \cII cut
    are both \mm points.
    Therefore, each pair of \mm points $p_1$ and $p_2$ represents a potential \cII cut if the line
    segment connecting them meets Condition ($ b $). This can be judged by checking
    if all pixels along $p_1 p_2$ are inside the shape.   %
    The number of \cII cuts must thus be less than $\binom{n}{2}$, with $n$ being the
    number of \mm points.

    The \CI cut
    has only one \mm endpoint.
    Thus, if we draw a line from an \mm to any
    arbitrary point on the shape contour,
    it can be considered as a potential \cI cut when meets Condition ($b$).
    The large number of possible cuts of this type makes the decomposition problem computationlly
    intractable. Next, we propose an approximate method.

    For an \mm point $p_i$ and a set of $n_d$ directions, we construct a set of $n_d$ potential part-cut hypotheses,
    each of which starts at $p_i$ and extends until it exits the shape.
    Let us suppose that a line intersects the contour at $p_i'$.
    Then $p_i p_i'$ is a \cI part-cut hypothesis.
	This approach is based on the observation that the human visual system does not work so accurately to differentiate part-cuts in very close directions.
	Therefore, as long as $n_d$ is sufficiently large
    (we find that $n_d=16$ suffices as shown in our experiments) and the directions are
    sampled evenly,
    the sampled $ n_d $ potential part-cut hypotheses can serve as a good approximation of the
    entire \cI part-cut set.
    The number of extracted \cI cuts should be less than $n_d \cdot n$ because
    there may not exist any line segment lying inside the shape in some directions.
    Recall that $n $ is the number of \mm  points on the contour
    of a shape.
    Moreover, if $p_i'$ is sufficiently close  to another \mm point
    (w.l.o.g., labelled as $ p_j $) along the contour,
    $p_ip_i'$ is then merged with $p_ip_j$ and regarded as a \cII cut.
    Here the ``sufficiently-close'' distance  is set to be less than  a  threshold
    equalling to  $  \min \{ 1\% \cdot m, d_{\rm min}  \} $. Here $ m $ is the
    length of the contour in pixels, and $ d_{\rm min} $ is the length of  the shortest edge in the
    polygon.
    In \cite{Psych9} the authors have defined a similar threshold.
    This threshold  allows for some noise in the segmentation data and at the same time
    rejects unlikely singularity points.

    We present the process of finding part-cut hypotheses in Algorithm~2. The output of this
    algorithm are two sets: $V$ and $H$, which contain the segmentation points and the part-cut
    hypotheses, respectively.

    \SetKwInput{KwInit}{Initialization}
    \SetVline
    \linesnumbered
    \begin{algorithm}[t]
    \caption{Part-cut hypotheses extraction.}
    \begin{algorithmic}
    	\KwIn{Shape $X$;}
    	\KwOut{The set of segmentation points (\ie, \mm points)  $V$, and the set of part-cut hypotheses $H$}
    	\KwInit {
    	$V \leftarrow \emptyset, H_1 \leftarrow \emptyset, H_2 \leftarrow \emptyset$;
    	}
    	Simplify the contour of $X$ by applying DCE, and store vertexes in $A$; \\
    	$V \leftarrow \{p \mid p \in A, \angle p > \pi+\delta \}$;\\
    	\For{$p_j \in V$}
    	{
    		$H_1 \leftarrow H_1 \cup \{ \text{\cI cuts starting at } p_i\}$; \\
    		\For{$p_j \in V$}
    		{
    			$H_2 \leftarrow H_2 \cup \{ p_i p_j \mid p_i p_j \text{ inside } X\}$;
    		}
    		$H \leftarrow H_1 \cup H_2$;
    	}
    \end{algorithmic}
    \label{alg:a2}
    \end{algorithm}

\subsection{Decomposing a shape into parts}

    After applying the two steps as described above, there are at most $ \binom{n}{2}+n_d \cdot n $
    part-cut hypotheses stored in $H$.
    The next step is to determine the ``true" part-cuts from the hypotheses in $ H $.

     Let us assign a boolean variable $ y_j \in \{ 0, 1\}$, ($ j = 1,\cdots, |H|  $),
     to each part-cut hypothesis in the set $ H $,
     where $ |H|$ is the size of the set $ H $.
     $ y_j = 1 $ means that the part-cut hypotheses with index $ j $ is identified as a final cut; and
     $ y_j = 0 $ indicates that the part-cut hypothesis $ j $ is discarded.
     In general, we can formulate an optimization problem as follows:
       \begin{align}
           \label{formula:f2}
                \min_{ y_1,\cdots, y_{|H|}  }
                \sum_j   \psi_u ( y_j  ) + \sum_{ jk  } \psi_v (y_j, y_k ).
        \end{align}
     The first unary term is the energy modelling the compatibility of data with label $ y_j $.
     The second term is the energy modelling the pairwise relationships.
     Since we do not have prior knowledge about
     the number of cuts for a particular shape,
     model selection criteria such as the Akaike information criterion
     may be used.
     Although it is possible to employ sophisticated optimization techniques
     such as integer programming to solve \eqref{formula:f2}, which is generally a
     NP-hard problem,
     we seek a sub-optimal solution using a greedy pruning method.

     Next we show how to properly define the two energy terms in \eqref{formula:f2}
     according to the short-cut rule and other psychophysics results, which is the core of our approach.

\subsubsection{Constraints from observations}
\label{sec:obs}

    Let us consider the unary energy term at first.
    The important fact is that a part-cut must satisfy some constraints that
    comply with visual perception.
    Here we formulate a few such fundamental constraints from observations
    which can be used to distinguish a part-cut from other candidates, mainly based on its local properties.
    It is to be highlighted that,
    due to the flexibility of our framework,
    it is easy to accommodate more constraints with minimal modification to the
    overall framework.

    \begin{figure}[t!]
    \begin{center}
    \includegraphics[width=0.4\textwidth]{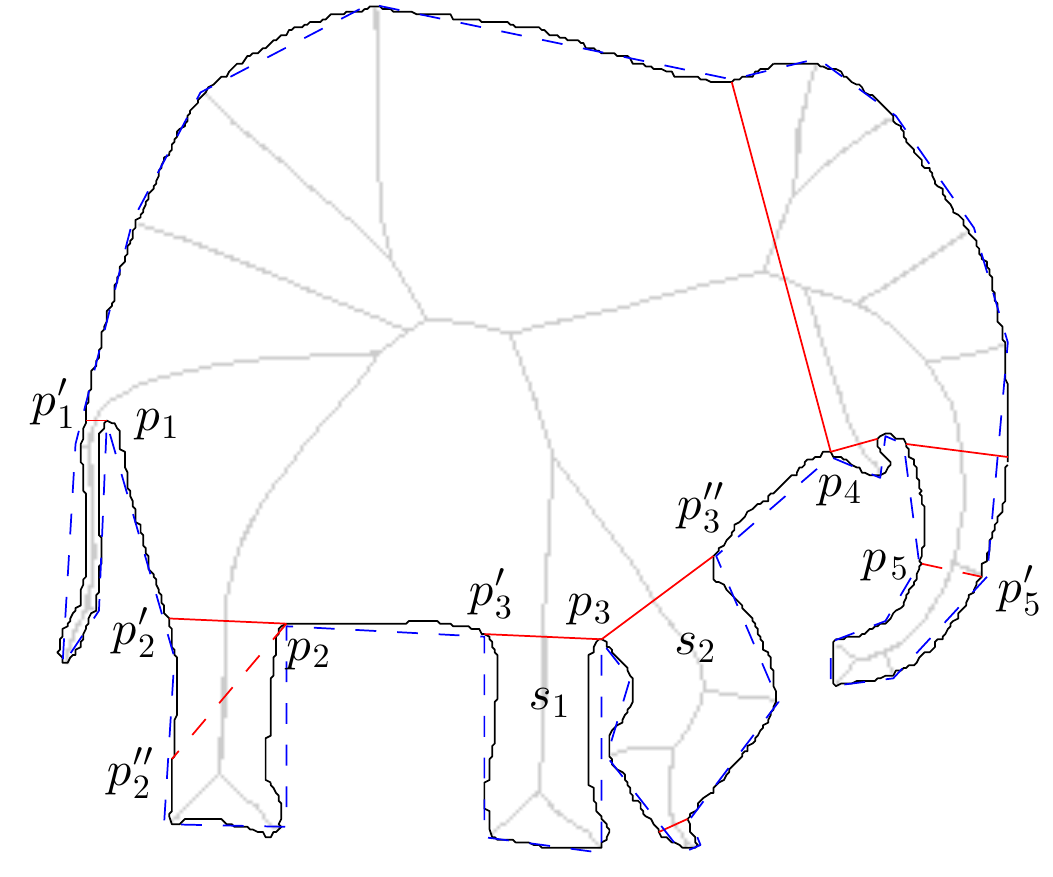}
    \end{center}
    \vspace{-.4cm}
       \caption{
       Illustration of observations on the elephant shape.
       The part-cuts shown in red solid lines all cross a skeleton (axis of local symmetry)
       almost orthogonally (Observation 1),
       and have at least one ``expanding" sides (Observation 3).
       Each \mm point can have two part-cuts at most, \eg, $p_3$ and $p_4$ both have two part-cuts (Observation 2).
       The \cI part-cut hypothesis $p_5p_5'$ is discard due to Observation 4.
       Skeletons of the shape are shown for interpretation, but they are never computed.
       }
    \label{fig:obs_elephant}
    \end{figure}

    \textbf{Observation 1}:  {\em A part-cut must cross an axis of local symmetry   `almost orthogonally'.}

    This observation is a strict version of Condition ($c$) of part-cuts as introduced in \cite{Hoffman1}.
    The requirement of `crossing almost orthogonally' is a result of the short-cut rule.
	Because the orthogonal ribs are usually the shortest cuts along the axis of local symmetry.
	See the toy example of $p_2p_2'$ and $p_2p_2''$ in Fig.~\ref{fig:obs_elephant}.

    \textbf{Observation 2}: {\em For every \mm point, there are at most two part-cuts going through it}.

    As illustrated in Fig.~\ref{fig:obs_elephant},
    $p_3$ may relate to at most two axes of local symmetry---$s_1$ corresponded to the left side of $p_3$ and $s_2$ corresponded to the right side.
    For each axis,
    only the shortest cut, which orthogonally crosses the axis,
    is kept according to Observation 1.

    \textbf{Observation 3}: {\em At least one side of a part-cut should be expanding}.

	The term ``expanding", whose counterpart is ``shrinking", means that departing from the part-cut in one side, local widths (the lengths of ribs along the axis of local symmetry) of the shape getting larger.
	Observation 3 is a complement of Observation 1.
    It is based on the fact that,
    if both sides are shrinking,
    the cut must be a local maximum of local width, which contradicts the short-cut rule\footnote{Refer Appendix \ref{sec:append_obs3} for more discussion about Observation 3.}.
    Clearly, if both sides of a part-cut are expanding, it is a neck or a neck-like \cI part-cut
    (Fig.~\ref{fig:crossAxis}(b) and (d)).
    Otherwise, it is a limb or a limb-like \cI part-cut (Fig.~\ref{fig:crossAxis}(a) and (c)).

    \textbf{Observation 4}: {\em A \cI part-cut should be salient}.

    This observation, which has its origin in \cite{ICCV5},
    is used to prevent noise on the outline of a shape.
    Unlike the \cII part-cut hypothesis, a \cI hypothesis has only one \mm end.
    If this end is caused by noise, it is less likely to be a conspicuous boundary of two parts.
    Therefore, we need to conduct further investigation to
    check the salience of a \cI hypothesis to be a part-cut.
    It can also distinguish bending from joint of two parts.
    See Fig. \ref{fig:obs_elephant} for example, $p_5p_5'$ is rejected since it is insufficient salient to decompose the trunk,
    while $p_1p_1'$ is accepted due to its strong segmentation effect in cutting the tail off from the body.

\subsubsection{Implementation of constraints}
\label{sec:obs_impl}
    Now with these observations, we can define the unary energy term in problem \eqref{formula:f2}.
    Given a particular part-cut hypothesis $ y_j $,
    if it violates any of the observations,
    its energy cost $ \psi_u (y_j)  $ is set to be very large.
    In other words,
    we set $ y_j = 0 $ and this part-cut hypothesis must be excluded.

    Here a problem is that
    all the observations except Observation 2 are qualitative descriptions
    instead of computable constraints.
    Here we propose to use the simplified shape polygon
    and the local information of the part-cut hypotheses to
    implement Observations 1, 3 and 4.

	For Observations 1 and 3, local symmetric axes are involved.
    Since the cost to compute the local symmetric axis is expensive,
    we use the simplified shape polygon to implement these two observations.
    In Fig. \ref{fig:crossAxis}, each part-cut hypothesis forms four angles with the polygon.
    For any hypothesis, the upside (downside) of which is expanding if and only if
    $\theta_1+\theta_4>\pi$ (or, $\theta_2+\theta_3>\pi$).
    Meanwhile, $\theta_1$ and $\theta_4$ (or, $\theta_2$ and $\theta_3$)
    are approximately equal per the requirment of Observation 1.
    Therefore, in practice,
    we  state that   a part-cut hypothesis satisfies Observations 1 and 3 if either
    $\theta_1,\theta_4 > \frac{\pi}{2}-\delta$ or
    $\theta_2,\theta_3 > \frac{\pi}{2}-\delta$ holds.
    Here $ \delta $ is a constant as defined in Section \ref{sec:211}.

	For Observation 4, the quantitative definition of  ``salience" is required.
	Mi and DeCarlo  \cite{ICCV5} discussed
    the salience of a part-cut hypothesis in detail.
    By borrowing their idea of \emph{transition strength}, here we construct the \emph{neighborhood histogram} for each \cI hypothesis
    in order to properly measure the salience of a \cI hypothesis.

\begin{figure}[t]
    \centering
        \includegraphics[width=0.5\textwidth]{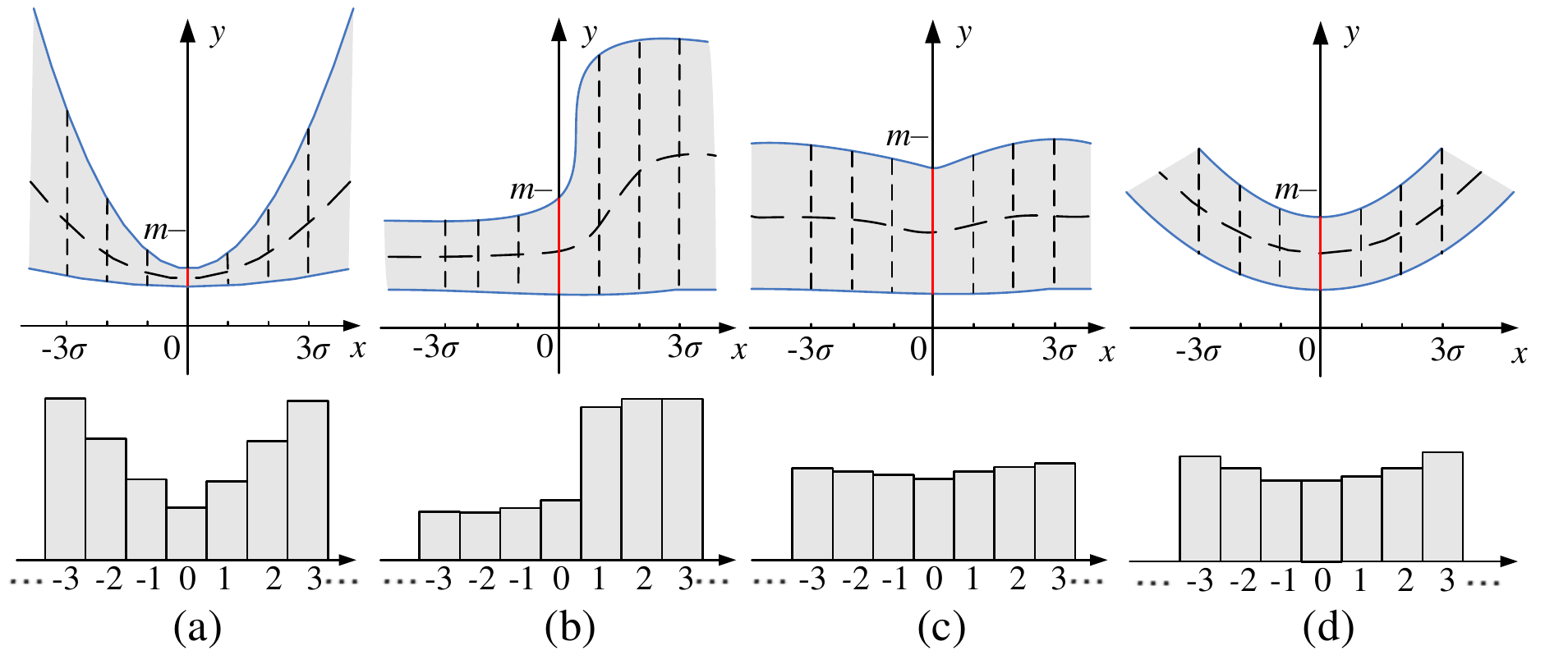}
      \vspace{-.6cm}
      \caption{
       Neighborhood histograms (bottom row)
       of \cI part-cut hypotheses (top row),
       where (a) and (b) change significantly while (c) and (d) are stable. }
    \label{fig:histogram}
    \end{figure}

    As shown in Fig.~\ref{fig:histogram},
    the $y$-axis in each plot is a \cI part-cut hypothesis.
    We draw a straight line in parallel to the $y$-axis for every $\hat{\sigma}$ pixels\footnote{For ease of calculation, we set $\hat{\sigma}=1$ all through the papar.}
    along the $x$-axis within a distance of $\pm{\sigma}$ pixels.
    Here $\sigma$ is the radius of neighborhood.
    Lengths of line segments inside the shape then construct the neighborhood histogram of the part-cut hypothesis,
    as shown at the bottom row of Fig.~\ref{fig:histogram}.
    Since a cut is orthogonal to the axis of local symmetry,
    these parallel line segments are also approximately orthogonal to the axis of local symmetry,
    based on the following two facts---the size of neighborhood $\sigma$ is small,
    and the axis of local symmetry is generally locally smooth \cite{1985symmetry}.
    In other words, they can approximately be seen as ribs of the axis.
    Thus the neighborhood histogram reveals the variation of the local width near the part-cut hypothesis.
    It is clear that the more significant the local width near a \cI part-cut hypothesis varies,
    the more salient it is.

    On the other hand, to fulfill the requirement of Observation 3,
    at least one side of its neighborhood histogram should  monotonically increase from the origin.
    In particular, if both sides increase as in Fig.~\ref{fig:histogram}(a),
    it is neck-like,
    and the more rapid they increase,
    the more salient it is.
    If only one side increases as in Fig.~\ref{fig:histogram}(b),
    it is limb-like,
    and the higher the two sides contrast,
    the more salient it is.
    We hereby use the statistics of the neighborhood histogram to measure the salience of a \cI hypothesis $h_j$.
    It is said to be salient if
    1) $\frac{\sigma_j}{\mu_j}>t_{h1},l_j<\mu_j$ or
    2) $\max\left\{
                        \frac{\mu_j^-}{\mu_j^+},\frac{\mu_j^+}{\mu_j^-}
            \right\}
            >
                t_{h2},l_j<\mu_j$,
    where $\mu_j$ and $\sigma_j$ are the
    mean and standard deviation of the entire histogram.
    $\mu_j^-$ and $\mu_j^+$ are means of the histogram's two halves.
    $t_{h1}$ and $t_{h2}$ are thresholds for salience.
    $l_j$ is the length of $h_j$ and $l_j<\mu_j$ is due to the requirement of
    the short-cut rule.
    The term $\max
                \left\{\frac{\mu_j^-}{\mu_j^+},\frac{\mu_j^+}{\mu_j^-}
                \right\}$,
    which measures the ``expansion" from one side of $h_j$ to the other side,
    is specially introduced for limb-like hypotheses.
    We regard that if the size expands two times from one side to the other,
    it could be consider as salient part-cut hypothesis,
    and thus set $t_{h2}=2$.
    The setting of $t_{h1}$ is discussed in the experiments (Section~\ref{sec:exp}).
    To support these statistic criteria,
    the radius of neighborhood $\sigma$ must be large enough.
    We find $\sigma = 5\hat{\sigma}$ is fine in experiments.

    Furthermore, we know that there are at most two part-cuts for every \mm point according to
Observation 2, and shorter candidates are more likely to be selected as part-cuts according to
the short-cut rule.
   To further reduce the number of part-cut hypothesis, we keep, at most, the
   shortest two double-minima part-cut hypotheses as well as
   the shortest two single-minimum part-cut hypotheses
   for each \mm points.
   This approximation seems not to compromise the accuracy,
   but reduces the number of part-cut hypotheses to examine in the next stage.
   Therefore, for each \mm points, we keep at most four part-cut  hypotheses as the candidates,
   from which at most two can be selected as true part-cuts due to Observation 2.
   Now, the number of part-cut hypotheses is no more than $ 4n $.

\subsubsection{Determination of part-cuts}
\label{sec:deter}

    The short-cut rule states that the cut length is a critical measure in deciding part-cuts.
    However, as shown in Fig.~\ref{fig:crossAxis},
    although shapes in each row share the same length of cuts and the same geometry on the left side,
    it appears that the strength of $p_ip_j$ partitioning the shape
    is stronger  than that of $p_ip_i'$ partitioning the shape.
    Mi and DeCarlo introduced the notion of \emph{transition
    strength} to describe this effect in \cite{ICCV5}.
    Their results suggest that
    the segmentation strength of a \cI cut is about half of a \cII cut.
    Therefore, we introduce
    the \emph{relative length} to encode  this difference.
    For a part-cut hypothesis $h_j$ with length
    $l_j$, its relative length
    $ \rl_j $ is defined as:
    \begin{equation}\label{formula:f3}
        \rl_j  = \left\{ {\begin{array}{*{20}l}
       {\frac{{l_j }}{R}}, & {{h_j\text{ is a \cII cut,}}}  \\
       {2\frac{{l_j }}{R}}, & {h_j{\text{ is a \cI cut,}}}  \\
    \end{array}} \right.
    \end{equation}
    where $R$ is the radius of the shape's minimum enclosing disk
    for normalization.

    The energy of a part-cut hypothesis $ h_j $ is a function of this relative
    length; \ie,
    \[
        \psi_u ( y_j  ) \equiv f ( \rl_j ) .
    \]
    Here $ y_j $ is the label of $ h_j $, and $ f( \cdot )$ is
    a monotonically increasing function since a shorter
    cut should carry lower energy and is always preferred.
    Since we use a greedy method that approximately solve the original NP-hard problem \eqref{formula:f2},
    we do not need to know the explicit form of $ f(\cdot)$. Instead we sort $h_j$ by values of $  f( \rl_j ) $,
    which can be simply achieved by sorting $ \rl_j $, $\forall j$.

	The term $ \psi_v ( y_j, y_k  ) $ in \eqref{formula:f2}
    models the pairwise relationship between $h_j$ and $h_k$.
    To ensure the visual parts are non-overlapped,
    the selected part-cuts should not intersect with each other.
    If two part-cut hypotheses intersect with each other, they are ``in conflict".
	There are some other specific cases of conflict between two part-cut hypothese.
    If two hypotheses $ h_j, h_k $ are in conflict, they cannot be selected simultaneously.
    Mathematically it means $ y_j \cdot y_k = 0  $.
    We can set the pairwise energy term $ \psi_v (  y_j = 1, y_k = 1 ) $ to be extremely large.
    If two hypotheses are compatible, the pairwise energy is set to zero.

    \SetKwInput{KwInit}{Initialization}
    \SetVline
    \linesnumbered
    \begin{algorithm}[t]
    \caption{Part-cut determination.}
    \begin{algorithmic}
    	\KwIn{Part-cut hypotheses $H$, segmentation points $V$;}
    	\KwOut{Part-cuts $C$;}
    	\KwInit {
    	$H' \leftarrow \emptyset, C \leftarrow \emptyset$;
    	}
    	$H \leftarrow \{h \mid h \in H, h \text{ satisfies Observations 1, 3, and 4} \}$; \\
    	\For{ $p_i \in V$}{
    		$H' \leftarrow H' \cup \{ \text{the shortest two part-cut hypotheses  }$ $\text{~~~~~~~~~~~~~~~~of each type starting at } p_i \text{ in } H\}$;
    	}
   		Sort part-cut hypotheses in $H'$ by their relative lengths in ascending order; \\
   		\For{$h_j' \in H'$}
   		{
   			\If{either endpoint of $h_j'$ has two part-cuts in $C$}{
   				continue \tcp*[r]{Observation 2}
   			}
   			\If{$h_j'$ does not conflict with any part-cut in $C$}{
   				$C \leftarrow C \cup \{ h_j' \}$;
   			}
   		}
    \end{algorithmic}
    \label{alg:a3}
    \end{algorithm}

    The part-cut determination process is illustrated in~Algorithm 3.
	Part-cut hypotheses are sorted by their relative lengths in
	ascending order.
	Those \cI ones are usually put in the rear of the examining queue due to their larger relative lengths.
    If the current part-cut hypothesis does not violate Observation 2
    and is not in conflict with any determined part-cuts,
    it is accepted.
    Part-cut hypotheses with shorter relative lengths are at  the front of the queue,
    therefore are preferred by the selection procedure.
    This can be viewed as the simplified implementation of the short-cut rule.

\subsection{Time complexity analysis}

    To analyze the time complexity of the proposed approach,
    our method must be divided into two stages---the
    simplification of the contour, and the determination of the part-cuts.

    The first stage is done using DCE \cite{DCE7},
    which has been shown to be $O(m\log m)$ in time
    complexity \cite{DCE17}.

    The most expensive computation in stage two is to judge whether a cut is
    inside the shape (the neighborhood histogram can be constructed at the same
    time) and to resolve conflicts between part-cut hypotheses. The former process
    consumes at most $O((\binom{n}{2} + n_dn)R)$ time when the number of cuts is less than
    $\binom{n}{2} + n_dn$ and the lengths of them are shorter than the radius of the shape's minimum enclosing disk $R$.
    The latter process needs $O((4n)(2n))$ time since the number of elements in $H^\prime$ and $C$ is less than $4n$ and $2n$, respectively.
    So the total time complexity of this stage is $O(n^2R)$.

    {
    Since $R$ is smaller than $m$, the total time complexity is $O(m\log m + n^2R + m)=O(m(n^2+\log m))$.
    }

    The work of \cite{convex12}, \cite{CVPR6}, \cite{ICCV5}, and \cite{2011ICCV_NearConvex} represents the most recent development for shape decomposition.
   The work of
   \cite{ICCV5} needs to identify SLS of the shape, which is $O(m^2)$.
    The most time-consuming part of \cite{CVPR6} and \cite{2011ICCV_NearConvex} is $O(t\log t)$, where $t$ is the number of pixels of the silhouette which is much larger than $m$ (in quadratic).
    As demonstrated in the next section, $n$ is always small (in dozens) in
    our experiments. Therefore the time complexity of our proposed method is lower than
    \cite{CVPR6,2011ICCV_NearConvex,ICCV5} in general.
    The work of
    \cite{convex12} is designed for approximate convex decomposition of polygons
and is very fast ($O(mr)$, where $r$ is the number of notches of the polygon).
    However, our method obtains more intuitive decomposition results than
\cite{convex12} as shown in the experiments.

    The average computational costs,
    on a standard dual-core desktop computer,
    of \cite{ICCV5}, \cite{CVPR6} and the proposed method in the experiments of Sec.~\ref{sec:ex_criteria}
    are 0.0048s, 10.89s and 0.1566s, respectively.
    Note that the experiments of \cite{ICCV5} used the well optimized C++ code provided by the author\footnote{http://masc.cs.gmu.edu/wiki/Software\#acd2d},
    while the other two are both  implemented in Matlab.
    The average decomposition time of \cite{2011ICCV_NearConvex} is 45.6s according to the authors.
    Those empirical computational results are consistent with the theoretical complexity analysis.

\section{Experimental Results} \label{sec:exp}

\subsection{Quantitative evaluation}
\label{sec:ex_criteria}

    Same as in image segmentation, evaluation of shape decomposition algorithms has been subjective,
    to an large extent.
    One usually has to judge the efficacy of a method by inspection of a small number of examples.
    Largely this is because shape decomposition itself is not a well-defined
    problem---one is not able to find a unique ground-truth decomposition of a shape,
    against which
    the output of an algorithm can possibly be compared. That may be the main reason
    why no quantitative evaluation was provided in shape decomposition work
    such as \cite{CVPR6,ICCV5,2011ICCV_NearConvex}.

    De Winter and Wagemans \cite{Psych9} have conducted
    a large-scale psychological study on
    how humans segment object shapes into parts
    based on a subset of Snodgrass and Vanderwart's everyday object data set (S~\&~V data set) \cite{shape16},
    which could help us to evaluate
    the decomposition results with human behavior.

\begin{figure}[t]
    \begin{center}
    \subfigure[]{\includegraphics[width=.095\textwidth]{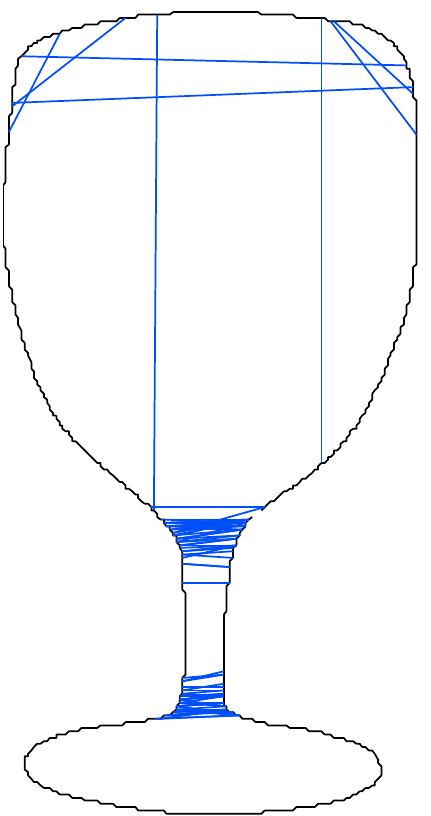}}
    \subfigure[]{\includegraphics[width=.112\textwidth]{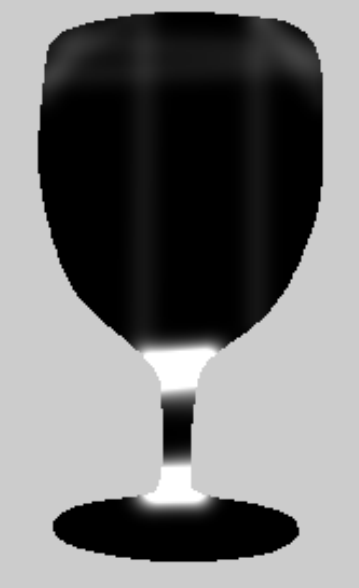}}
    \subfigure[]{\includegraphics[width=.095\textwidth]{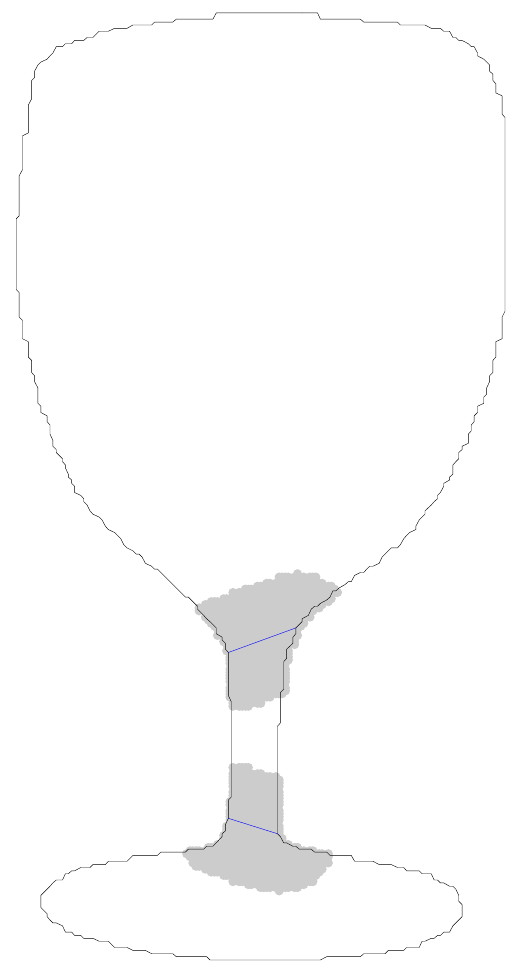}}
    \subfigure[]{\includegraphics[width=.117\textwidth]{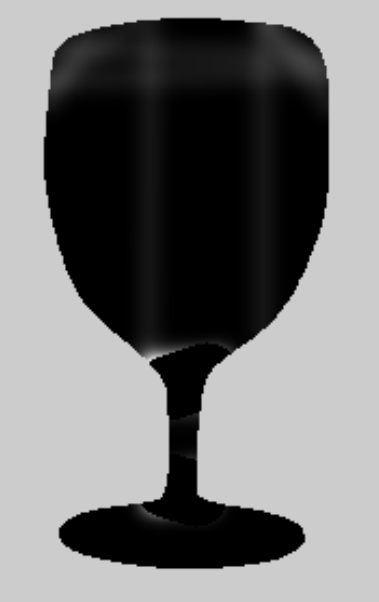}}
    \end{center}
    \vspace{-.4cm}
    \caption{
       (a) Segmentation lines drawn by subjects in De Winter and Wagemans' experiments.
       (b) Segmentation density map.
       (c) Part-cuts and their masking area.
       (d) Masked segmentation density map.
    }
    \label{fig:psychology}
    \end{figure}

\begin{figure}[t]
    \begin{center}
    \includegraphics[width=.4\textwidth]{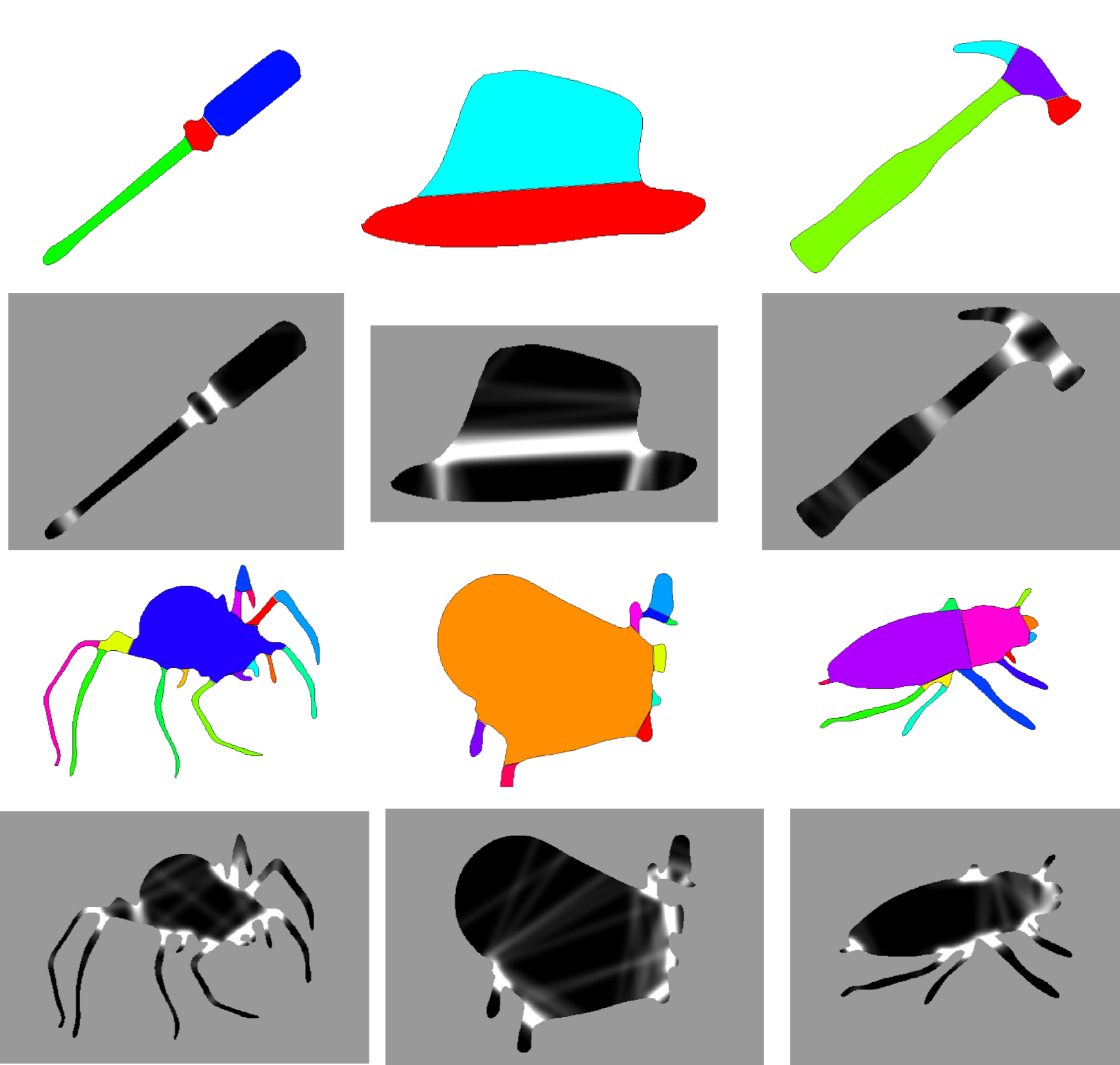}
    \end{center}
    \vspace{-.3cm}
    \caption{
       Decomposition of some shapes from S~\&~V dataset (pseudocolor image),
       and their segmentation density maps (grey-scale image).
    }
    \label{fig:s_v}
    \end{figure}

    S~\&~V data set consists of 260 line drawing of everyday objects,
    88 of which were selected and converted into outline shapes
    as the stimulus materials to be decomposed by 201 subjects (first-year university students).
    Each subject is asked to segment a set of figures into parts by drawing lines
    (either straight or curved but straight lines were reported in \cite{Psych9}
    when most of the drawn lines were straight).
    One example of drawn lines from all subjects on a ``glass" is shown in Fig.~\ref{fig:psychology}(a).
    We see that most subjects share the same opinion to separate the glass around the two joints of the stem,
    while a few others hold different views.
    To integrate opinions from all subjects,
    we propose to construct the \emph{segmentation density map}
    for every shape with a Gaussian mixture like model.
    The segmentation density in any pixel $x_i$ of a shape is the accumulation of all drawn lines:
    \begin{equation}\label{formula:density}
    q(x_i ) = \sum\limits_j {q_j (x_i ) =
    \sum\limits_j {\exp (\frac{{ - d_j^2 (x_i )}}{{\sigma ^2 }})} },
    \end{equation}
    where $d_j(x_i)$ is the distance from $x_i$ to the $j$th drawn line,
    $\sigma$ is the radius to construct neighborhood histogram in Sec.~\ref{sec:obs_impl}.
	Fig.~\ref{fig:s_v} presents some shapes  decomposed by the proposed method
	and their segmentation density maps from S~\&~V dataset.
	It is shown that most part-cuts lie on high density areas.
	Next we propose a criterion to evaluate the quantitative performance of a decomposition algorithm.

    Given a set of part-cuts $C$ on the shape, we say a pixel is masked if it is within the distance of $3\sigma$ (due to the well-kown ``Three-sigma rule") to any part-cut.
    Let
    \begin{equation}\label{formula:mask}
    \begin{array}{l}
     \mu _{{\rm{masked}}}  = \mathop {{\rm{mean }}}\limits_{\exists j,D_j (x_i ) \le 3\sigma } q(x_i ):\mathop {\text{mean}}\limits_{x_i } q(x_i ), \\
     \mu _{{\rm{unmasked}}}  = \mathop {{\rm{mean }}}\limits_{\forall j,D_j (x_i ) > 3\sigma}
     q(x_i ):\mathop {\text{mean}}\limits_{x_i } q(x_i ), \\
     \end{array}
    \end{equation}
    where $D_j(x_i)$ is the distance from $x_i$ to the $j$th part-cut.
    The larger $\mu_\text{masked}$ is,
    the better $C$ is compatible with the drawn lines;
    and the smaller $\mu_\text{unmasked}$ is,
    the more drawn lines $C$ covers.
    That is, $\mathcal{H}=\mu_\text{masked}:\mu_\text{unmasked}$ represents the overall similarity between $C$ and the experimental results of \cite{Psych9} (higher is better).
    However, a large number of part-cuts,
    covering a major proportion of the shape,
    may lead to a very small $\mu_\text{unmasked}$.
    Then $\mathcal{H}$ is significantly high
    even when $C$ is not quite compatible with the drawn lines.
    So, as a precondition, the number of part-cuts $|C|$
    should be roughly consistent with the psychophysical results,
    which is $3.97$ per shape
    (22 shapes and $68.6$ drawn lines per subject with 21.4\% ``error" lines uncounted).

    We compare the performance of three methods by average values on S~\&~V data set in Table \ref{tab:comparation}.
    It is clear that the proposed method is more accordant with the experimental results of \cite{Psych9}.
    Surprisingly, the method in \cite{CVPR6} performs much worse than the others.
    One possible reason may be the absence of visual naturalness in its definition of concavity.

    \begin{table}[t]
    \renewcommand{\arraystretch}{1.23}
    \caption{Comparison of decomposition results on S~\&~V data set. $\overline{\mathcal{H}}$ represents the overall similarity between $C$ and human decomposition. Higher is better.} %
    \label{tab:comparation}
    \centering
    \begin{tabular}{ r |c|c|c|c}
    \hline
    Method & $\overline{|C|}$ & $\overline{\mu_\text{masked}}$ & $\overline{\mu_\text{unmasked}}$ & $\overline{\mathcal{H}}$ \\ %
    \hline %
    ACD \cite{convex12} & 4.18 & 3.49 & 0.69 & 6.85  \\ \hline %
    CSD \cite{CVPR6}    & 3.80 & 3.09 & 0.78 & 4.72  \\ \hline %
    Ours                & 4.07 & 3.77 & 0.66 & 8.54 \\ %
    \hline
    \end{tabular}
    \end{table}

\subsection{Evaluation of parameters}

    \begin{figure}[t]
    \begin{center}
    \subfigure[]{\includegraphics[width=.2\textwidth]{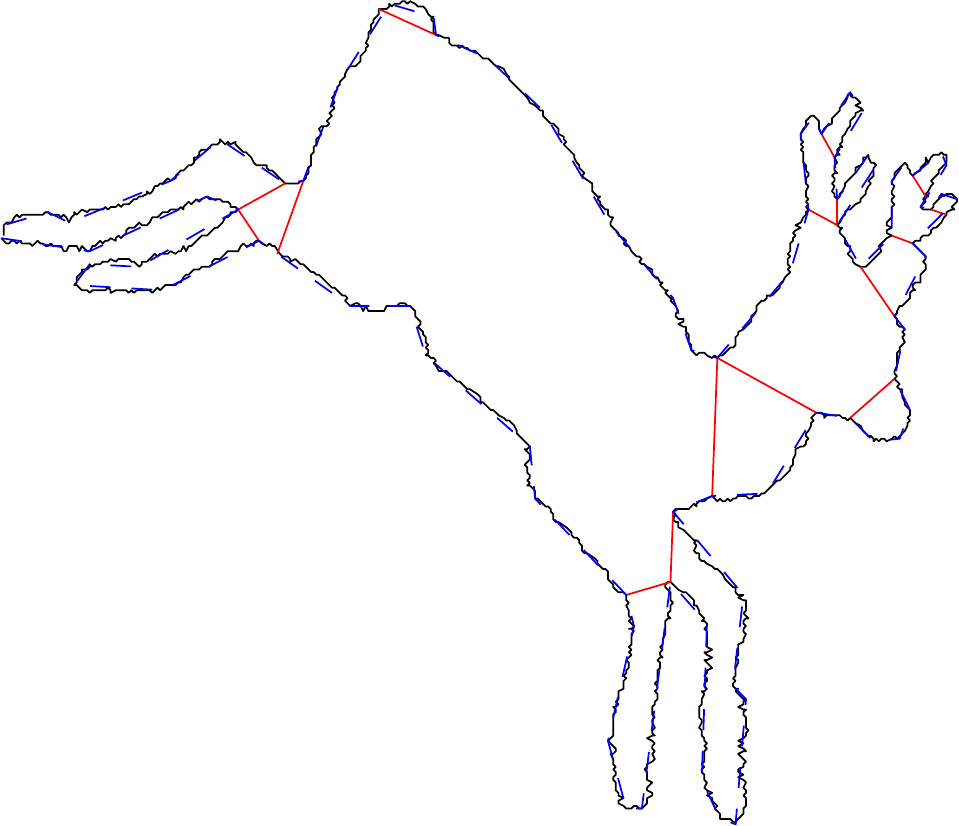}}
    \subfigure[]{\includegraphics[width=.2\textwidth]{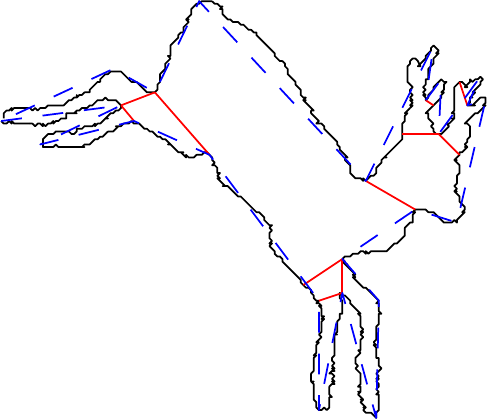}} \\
    \vspace{-.2cm}
    \subfigure[]{\includegraphics[width=.2\textwidth]{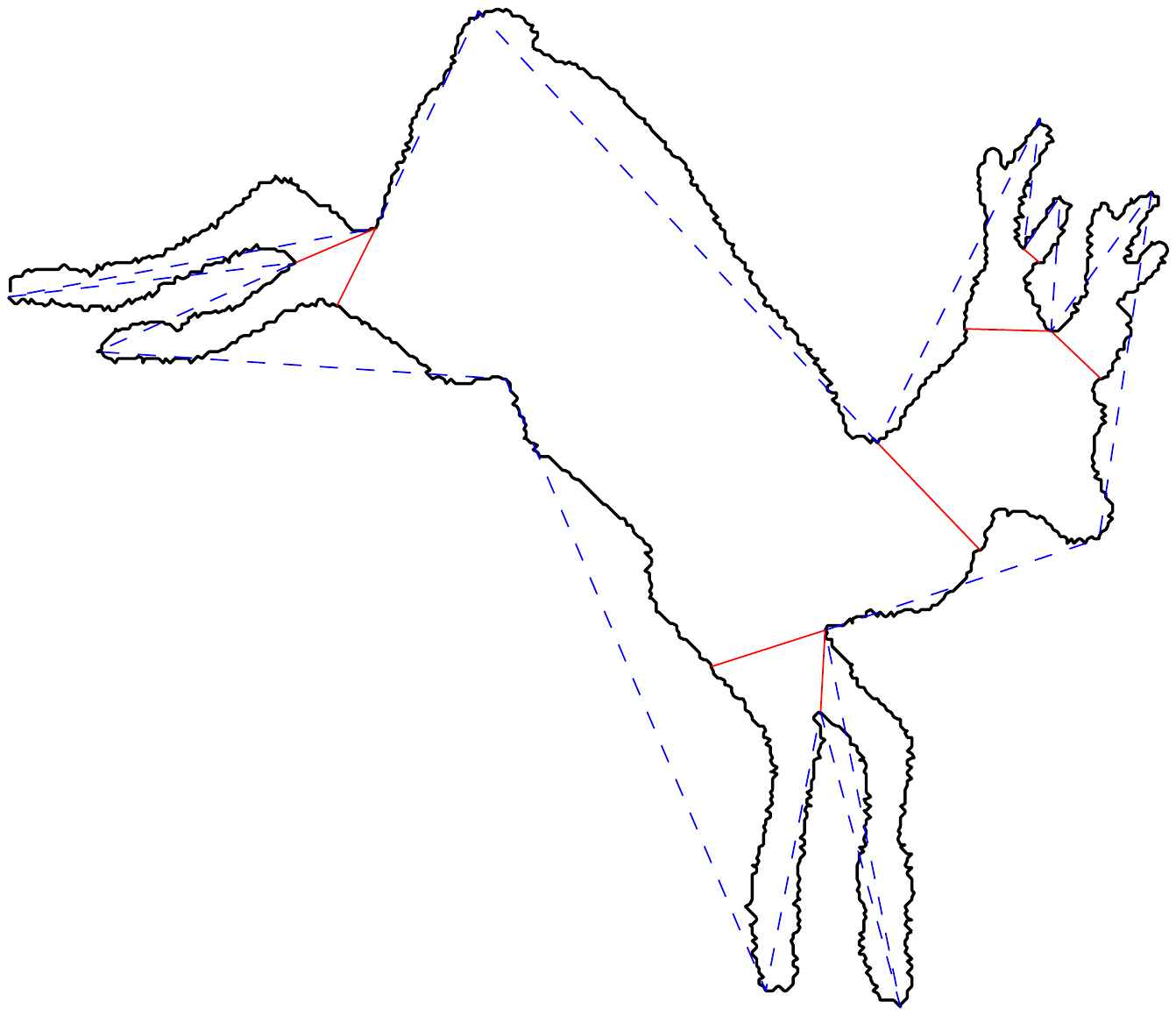}}
    \subfigure[]{\includegraphics[width=.2\textwidth]{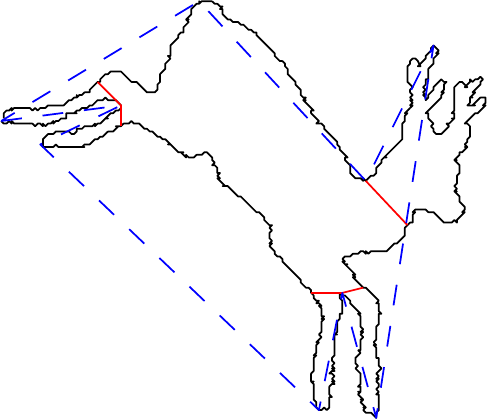}}
    \end{center}
    \vspace{-.4cm}
    \caption{
        The decomposition results by the proposed method, with
        (a) $t_{\text{DCE}}=0.1$,
        (b) $t_{\text{DCE}}=0.5$,
        (c) $t_{\text{DCE}}=1$ and
        (d) $t_{\text{DCE}}=3$, respectively.
        The simplified polygons are in blue dashed lines while the determined part-cuts are in red solid lines.
    }
    \label{fig:deer}
    \end{figure}

    A number of parameters have been introduced in our algorithm,
    which is a caveat of the proposed method.
    This is mainly because that human visual perception is a complicated process involving a lot of different (even conflicting, sometimes) principles \cite{Hoffman1}, which can hardly be explained by several simple formulas.
    Nevertheless, most parameters possess clearly defined perceptual meanings and have been discussed accordingly when they are introduced.
    Other parameters include the stopping parameter $t_{\text{DCE}}$ of DCE,
    the number of directions $n_d$ for generating \cI part-cut hypotheses,
    and the threshold $t_{h1}$ associated with the neighborhood histogram.

    The parameter $t_{\text{DCE}}$ tells how similar the simplified polygon with the origin shape boundary.
    Most discussions in Section \ref{sec:main} are based on the assumption that
    the polygon obtained by DCE is an approximate version of the shape's boundary.
    Thus, $t_{\text{DCE}}$ should be small to maintain a high degree of similarity.
    We examine the impact of this parameter on the final performance of our method.
    As shown in Fig.~\ref{fig:deer}, the proposed method works well for different values of $t_{\text{DCE}}$.
    With a small $t_{\text{DCE}}$, the detail of the shape boundary is kept,
    which in general introduces a large number of small parts. %
    When the value of $t_{\text{DCE}}$ increases,
    the decomposition tends to miss more detail parts and tolerate more distortions at the same time.

\begin{figure*}[t]
    \begin{center}
    \subfigure[ACD \cite{convex12}]{\includegraphics[width=.3\textwidth]{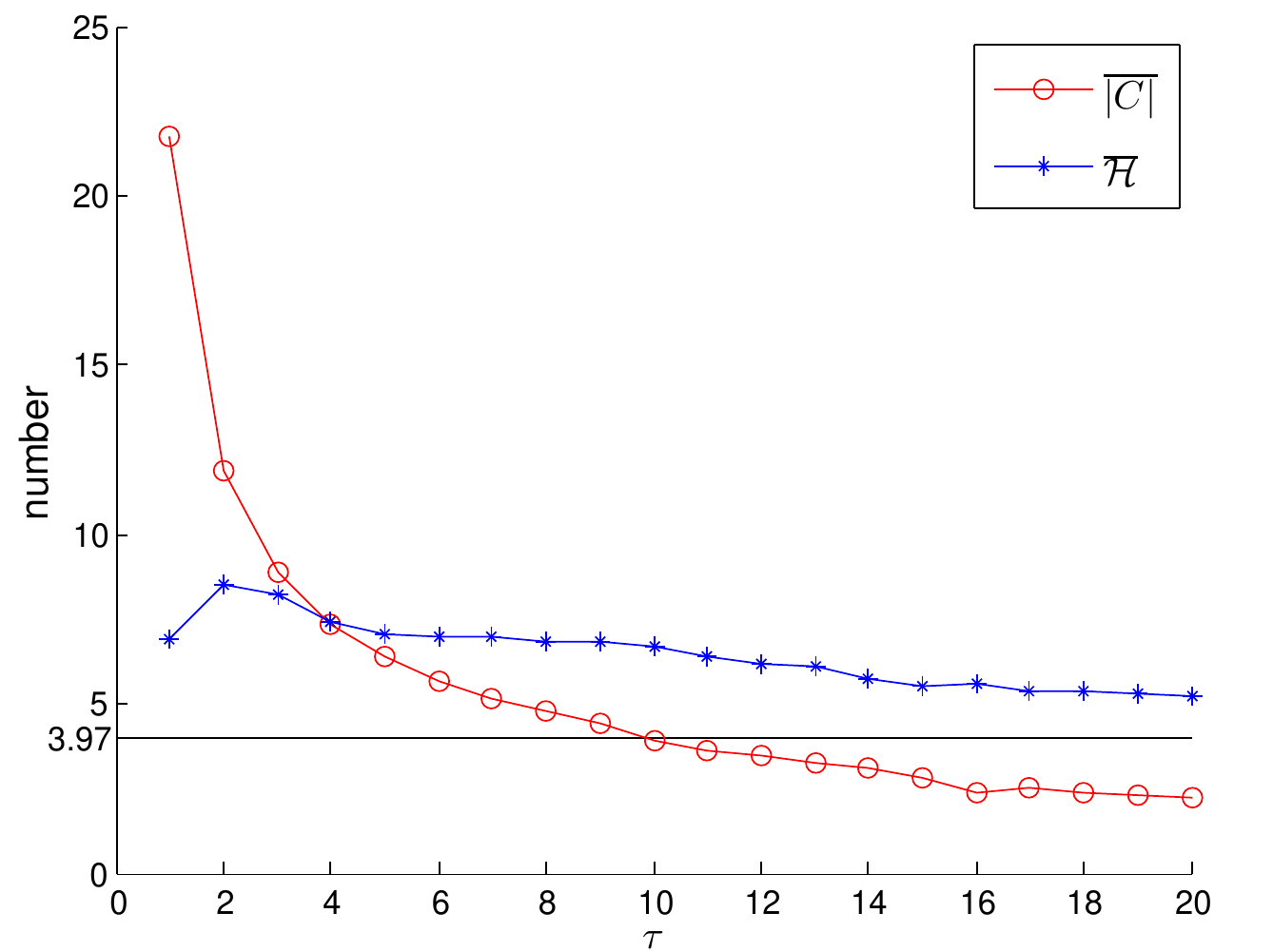}}
    \subfigure[CSD \cite{CVPR6}]{\includegraphics[width=.3\textwidth]{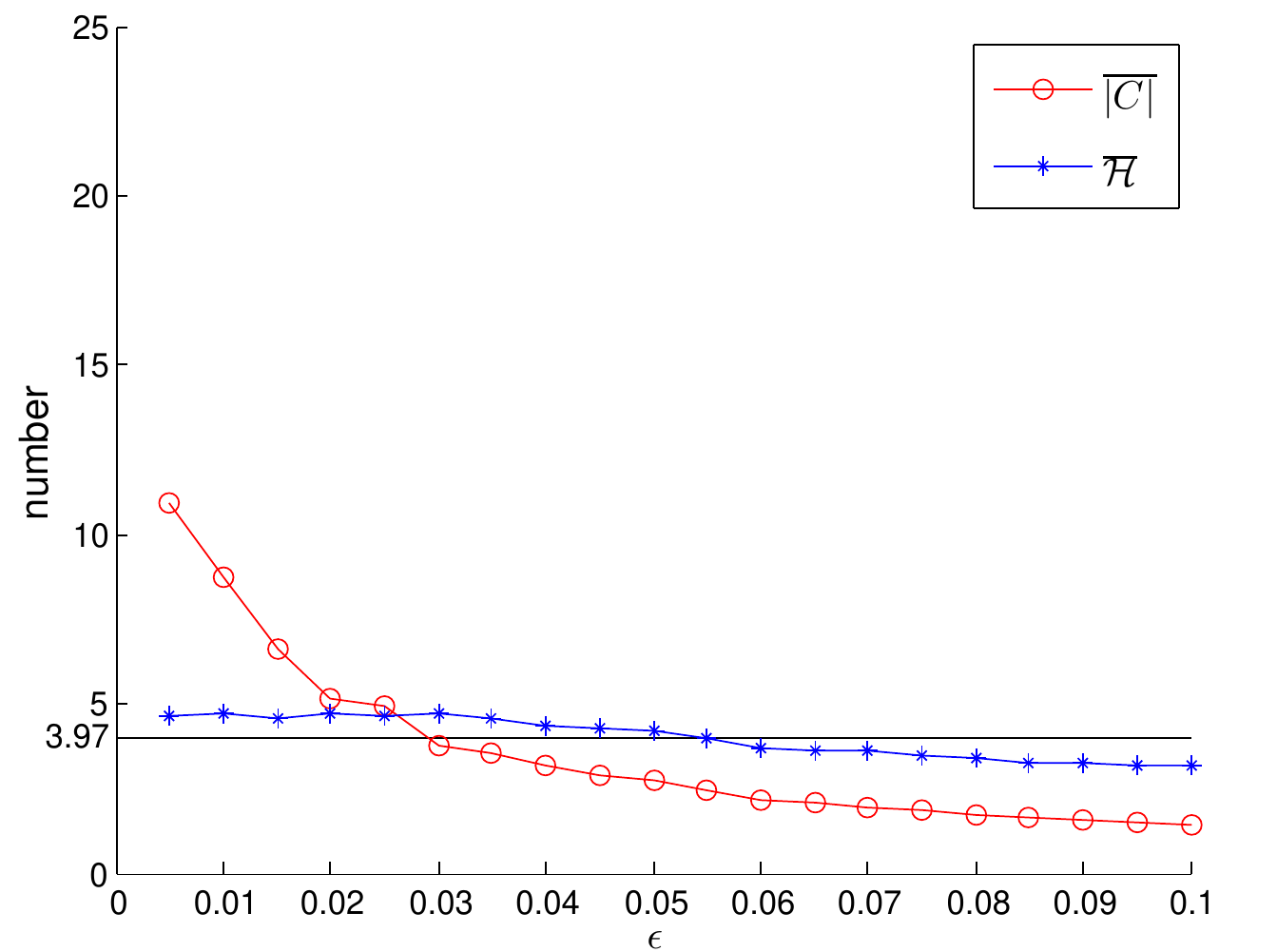}}
    \subfigure[Ours]{\includegraphics[width=.3\textwidth]{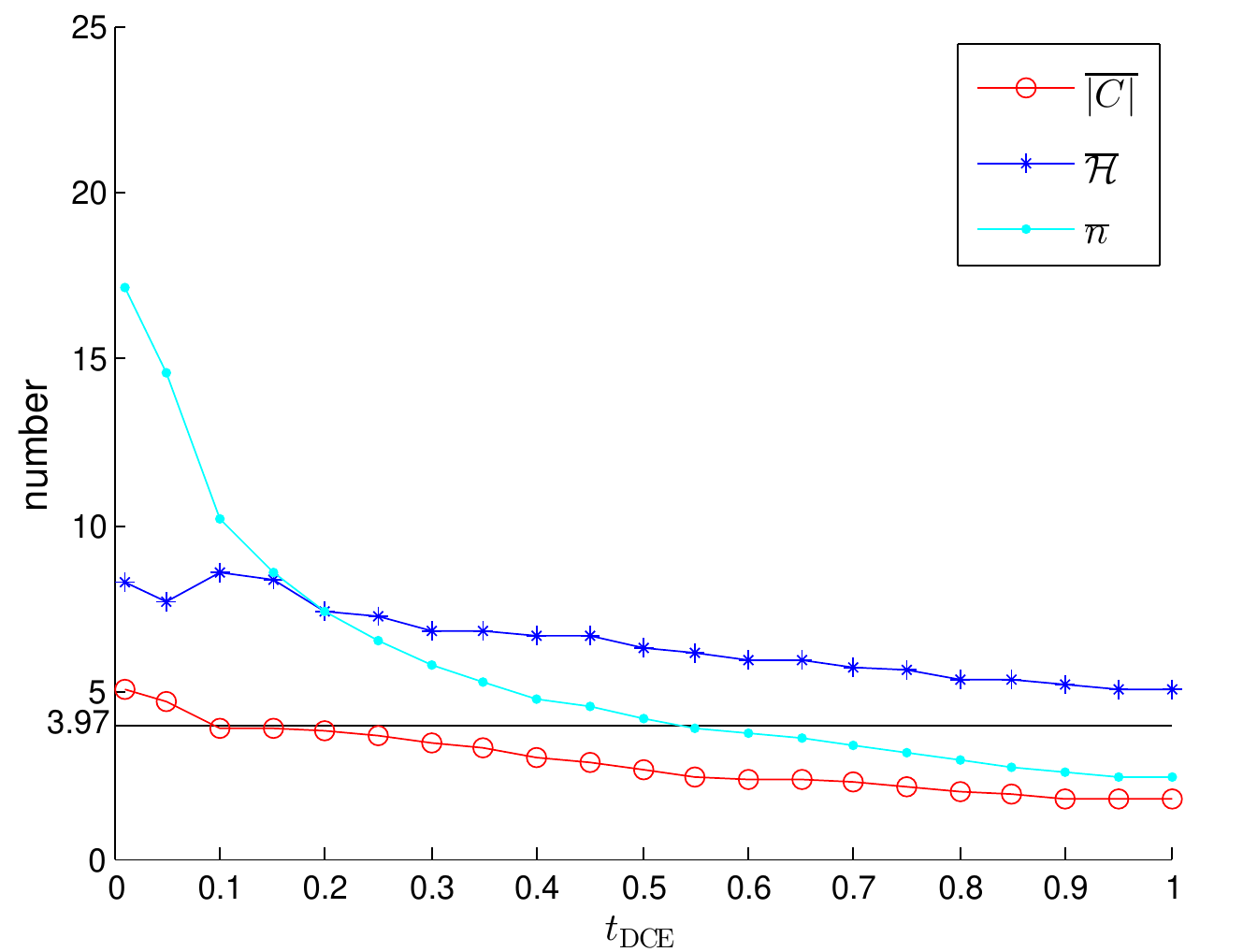}}
    \end{center}
    \vspace{-.4cm}
    \caption{
    Impact of the convex parameters for the three algorithms on the S \& V data set.
    }
    \label{fig:DCE}
    \end{figure*}

	Fig. \ref{fig:DCE}(c) summaries the impact of $t_{\text{DCE}}$ on
	the performance on the S~\&~V data set.
	The average number of part-cuts
	$\overline{|C|}$ is always not far from the psychophysical result of 3.97.
	The highest $\overline{\mathcal{H}}$ is obtained (with $t_{\text{DCE}}$ around 0.1)
	when $\overline{|C|}$ approximately fits it.
	It also shows that the average number of \mm points $\overline{n}$ is always small (less than 20),
	which guarantees the low complexity of the proposed algorithm.

    For comparison, we also plot the influence of $\tau$ to ACD and $\epsilon$ to CSD
    ($\tau$ and $\epsilon$ are both thresholds for concavity similar to $t_{\text{DCE}}$)
    in Fig. \ref{fig:DCE} (a) and (b), respectively.
    In (a), $\overline{|C|}$ is very large at a small $\tau$
    and decreases almost exponentially when $\tau$ increases.
    The highest $\overline{\mathcal{H}}$ is obtained when $\overline{|C|}$ is
    three times larger than the psychophysical results.
    It is lower when $\overline{|C|}$ reaches 3.97 with $\tau$ being around 10.
    In (b), $\overline{\mathcal{H}}$ keeps lower than 5,
    and $\overline{|C|}$ reaches 3.97 with $\epsilon$ being around 0.03.

    We also evaluate the influence of the other two parameters $n_d$
    and $t_{h1}$ on the S~\&~V data set.
    In the experiments, $n_d$ varies
    from 8 to 32 with an increase  of 8 at each step and $t_{h1}$
    ranges from 0.2 to 0.8 with an increase of 0.2 at each step.

    The results are reported in Table \ref{tab:parameter}.
    For $\overline{\mathcal{H}}$, the higher is better,
    and for $\overline{|C|}$, the closer to 3.97 is better.
    The best parameter settings are $n_d=8$ and $t_{h1}=0.4$.
    We can see that $n_d=16$ is usually sufficient
    for generating \cI part-cut hypotheses.
    When $n_d>16$, not only the complexity increases,
    but the decomposition results are also  less consistent with  the psychological results.

    \begin{table}[t]
    \renewcommand{\arraystretch}{1.23}
    \caption{
        The score of $\overline{\mathcal{H}}$ (left) and
        $\overline{|C|}$ (right) for the S \& V data set based on
        different pairs of parameters.
        }
    \label{tab:parameter}
    \centering
    \begin{tabular}{ c|c|c|c|c }
    \hline
     \backslashbox{$t_{h1}$}{$n_d$}& $8$ & $16$ & $24$ & $32$ \\
    \hline
    $0.2$ & 8.48 / 4.23 & 8.44 / 4.51 & 8.40 / 4.61 & 8.51 / 4.82  \\
    \hline
    $0.4$ & 8.59 / 3.93 & 8.54 / 4.07 & 8.59 / 4.23 & 8.35 / 4.32  \\
    \hline
    $0.6$ & 8.33 / 3.86 & 8.35 / 3.95 & 8.34 / 4.08 & 8.10 / 4.18 \\
    \hline
    $0.8$ & 8.33 / 3.78 & 8.28 / 3.91 & 8.24 / 3.98 & 8.01 / 4.10 \\
    \hline
    \end{tabular}
    \end{table}

\subsection{More results}

\begin{figure}[t]
    \begin{center}
    \includegraphics[width=.48\textwidth]{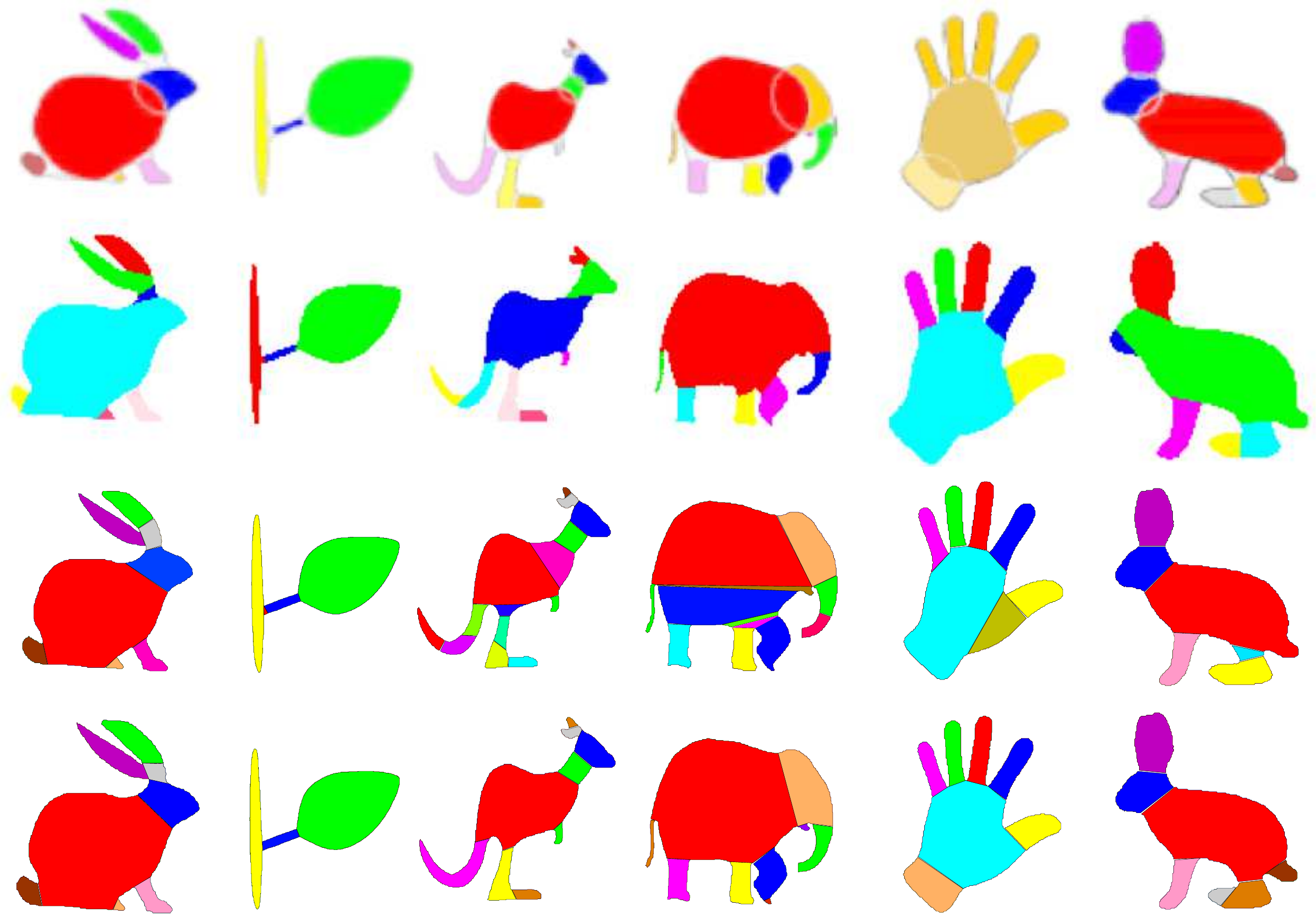}
    \end{center}
    \vspace{-.2cm}
    \caption{
        From top to bottom: decomposition results of \cite{ICCV5}, \cite{CVPR6}, \cite{convex12} and our method.
     }
    \label{fig:compare}
    \end{figure}

    To further evaluate the visual naturalness of the proposed
    algorithm, we compare the decomposition results of \cite{ICCV5},
    \cite{CVPR6}, \cite{convex12} and our method in Fig.~\ref{fig:compare}.
    As we can see, the first and the fourth row produce similar and intuitive results,
    while the second and the third row may parse a long bend (\eg, the tail of the kangaroo) into parts.

\begin{figure}[t]
    \begin{center}
    \includegraphics[width=.48\textwidth]{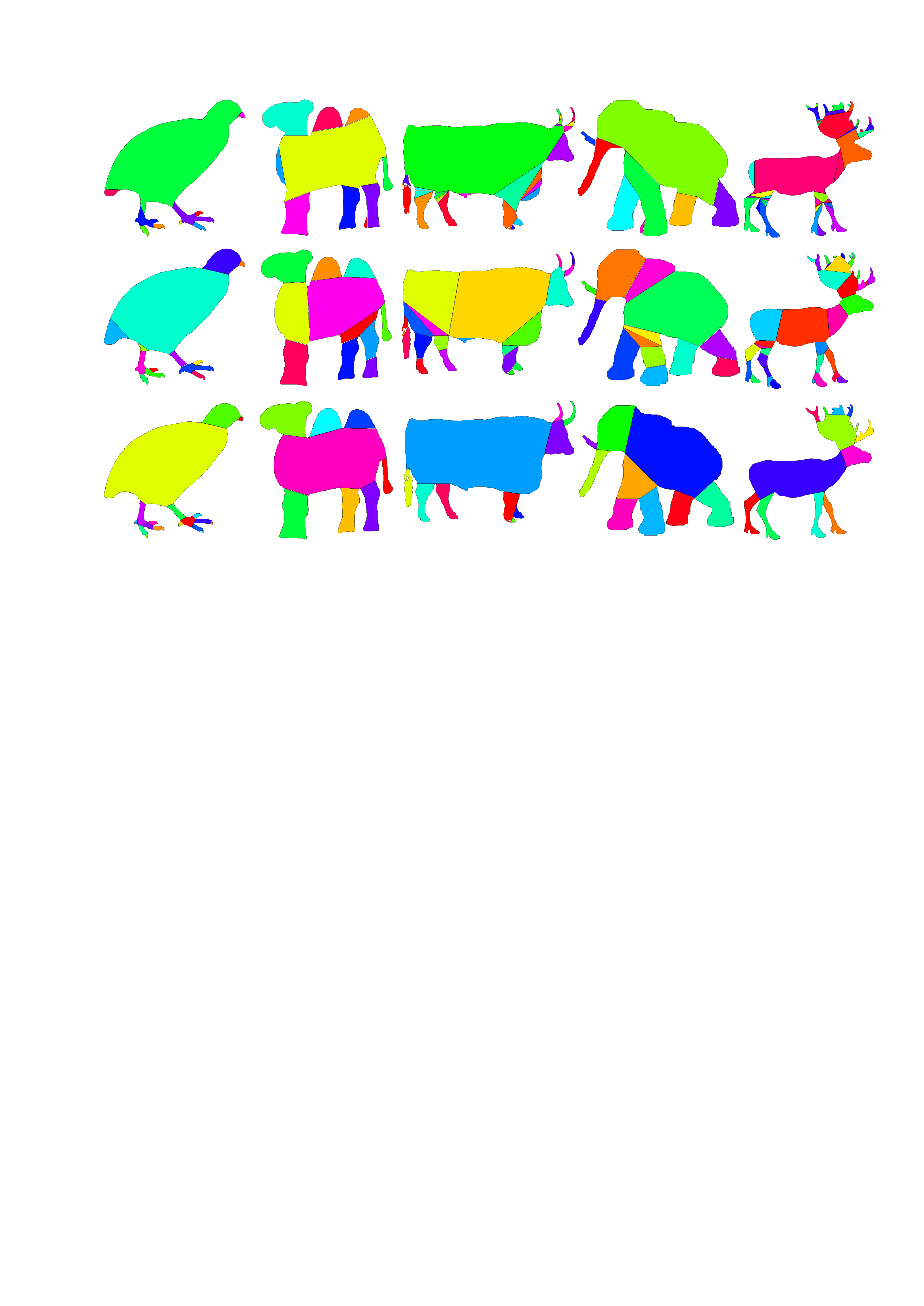}
    \end{center}
    \vspace{-.2cm}
    \caption{
        From top to bottom: decomposition results of some shapes from MPEG-7 shape database by \cite{CVPR6}, \cite{convex12} and our method.
     }
    \label{fig:mpeg}
    \end{figure}

    Fig.\ \ref{fig:mpeg} compares the decomposition results of some shapes from the
    MPEG-7 shape database produced
    by ACD \cite{convex12}, CSD \cite{CVPR6} and our method.
    It can be seen that our method produces less part-cuts and the results are more natural.

\begin{figure}[t]
    \begin{center}
	\subfigure[]{\includegraphics[height=.16\textheight]{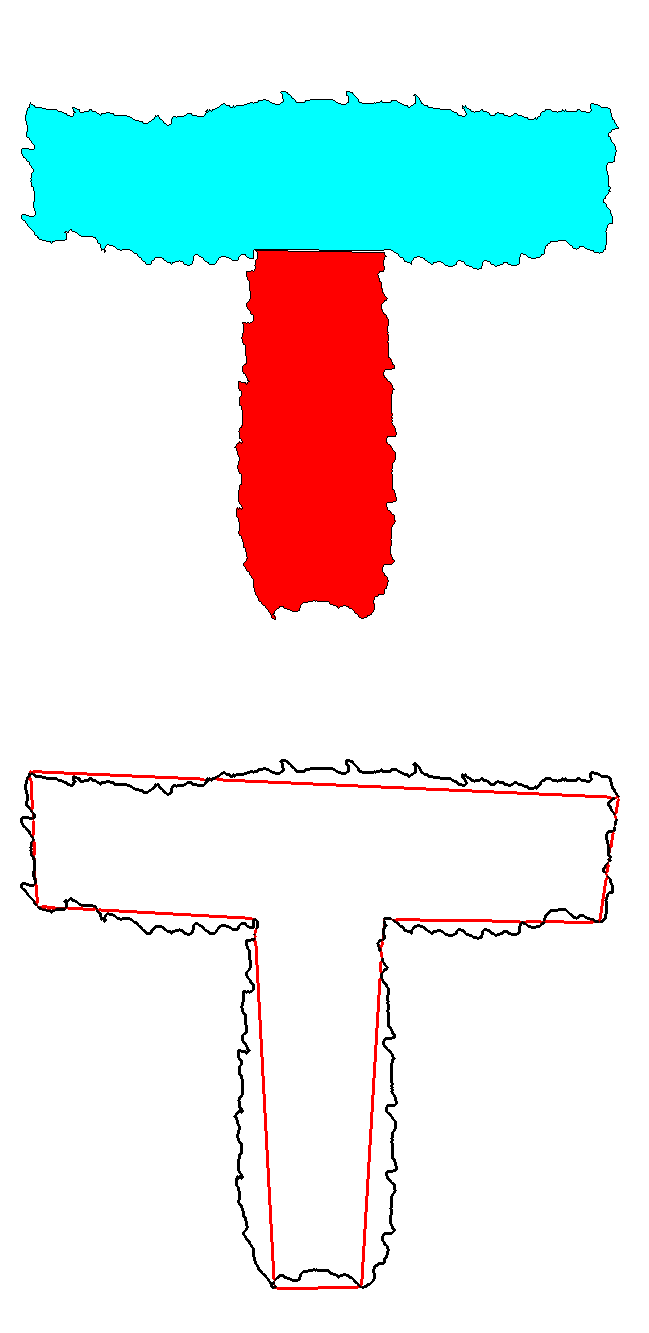}}
	\subfigure[]{\includegraphics[height=.16\textheight]{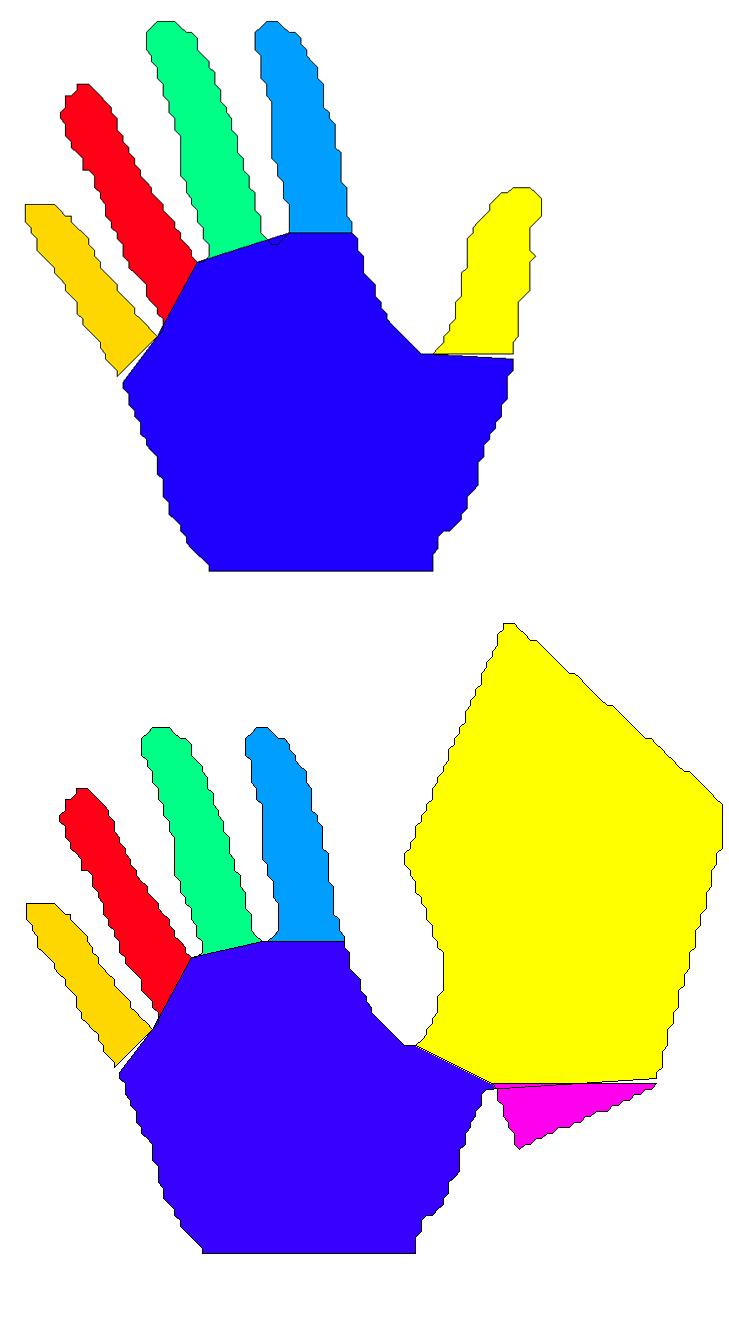}}
	\subfigure[]{\includegraphics[height=.16\textheight]{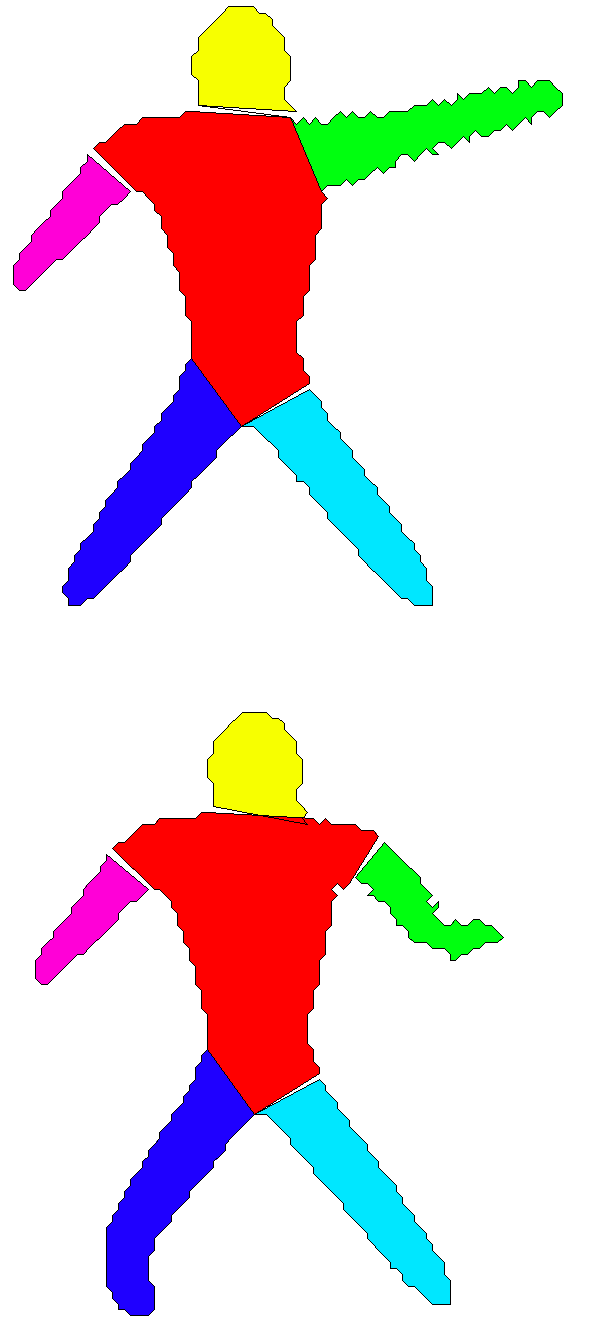}}
	\subfigure[]{\includegraphics[height=.16\textheight]{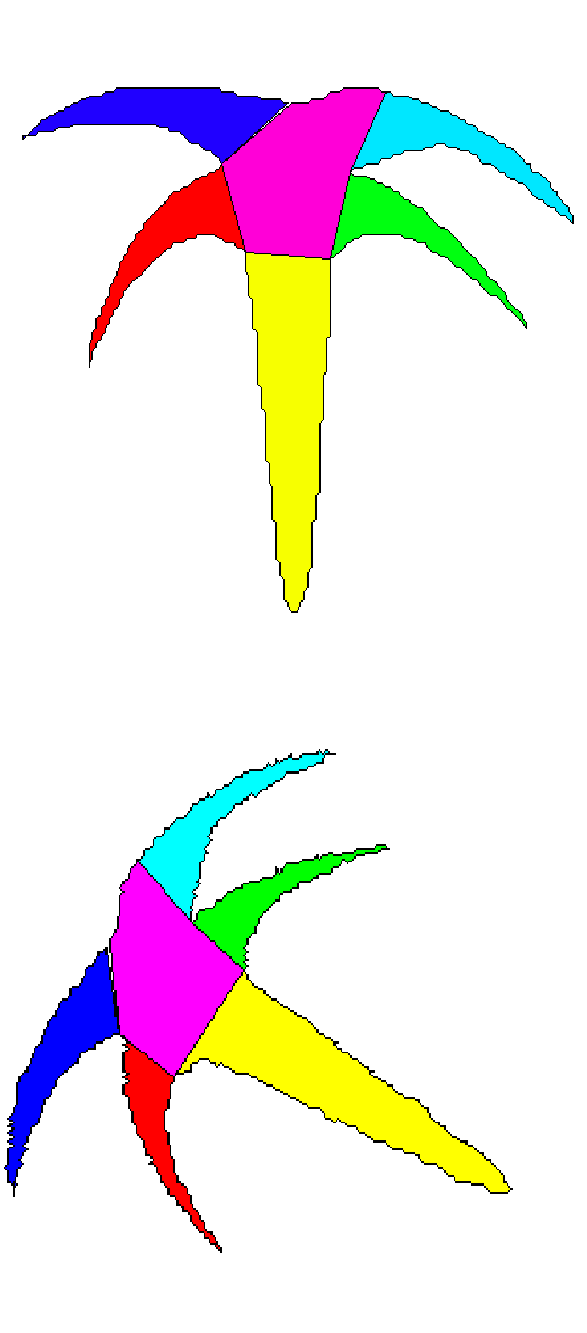}}
    \end{center}
    \vspace{-.4cm}
    \caption{
           Decomposition examples in the presence of (a) noise, (b) occlusion, (c) articulation and (d) rotation.}
    \label{fig:robust}
    \end{figure}

    Fig.~\ref{fig:robust} demonstrates the robustness of our method in the presence
    of noise, occlusion, articulation and rotation.
    We deal with noise by increasing $t_\text{DCE}$. As in the first column, the noised ``T" shape is firstly de-noised to a closed polygon (drawn in red lines) and then decomposed into two parts.
    We also see that occlusion in the second column does not affect the decomposition process in the un-occluded part of the shape. In the third column, the left hand of the man in the right column moves around his shoulder. Our method separates it from his body in both cases.
    In the last column, we can see that the
    angle of the single-minimum part-cut alters slightly.
    This is because we only generate single-minimum part-cut hypotheses
    in $n_d$ directions. However, this alteration does not have a significant impact
    on the performance when $n_d$ is set to a sufficiently large value.

\subsection{Hand gesture recognition}

\begin{figure}[t]
    \begin{center}
    \subfigure[]{\includegraphics[width=.12\textwidth]{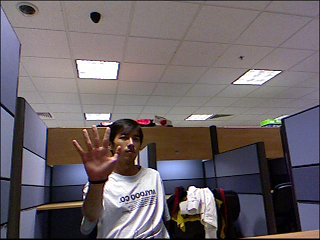}}
    \subfigure[]{\includegraphics[width=.12\textwidth]{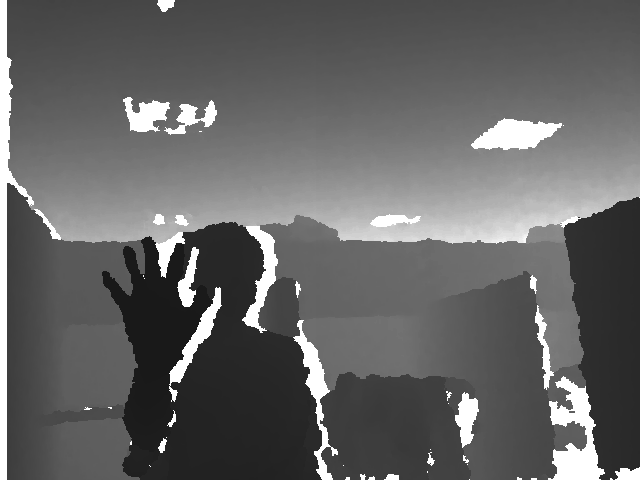}}
    \subfigure[]{\includegraphics[width=.12\textwidth]{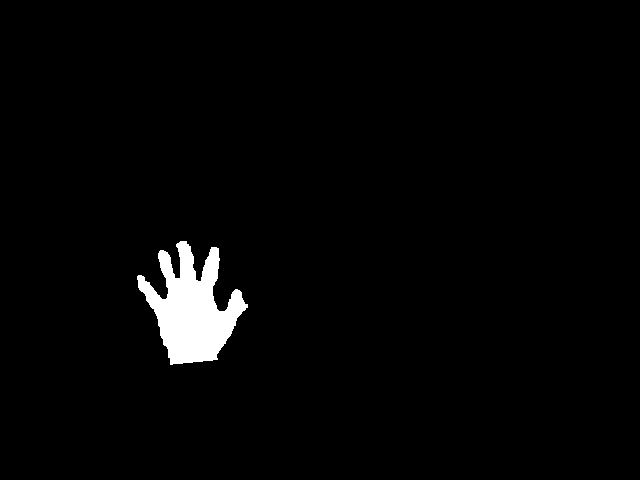}}
    \subfigure[]{\includegraphics[width=.0912\textwidth]{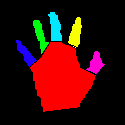}}
    \end{center}
    \vspace{-.4cm}
       \caption{
       Illustration of the hand gesture recognition experiment. From left to right: the color image, the depth image, the extracted hand shapes and the decomposition result.
       }
    \label{fig:handGesture}
    \end{figure}

\begin{figure}[t]
    \begin{center}
    \includegraphics[width=.08\textwidth]{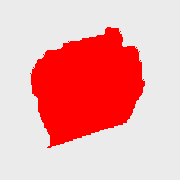}
    \includegraphics[width=.08\textwidth]{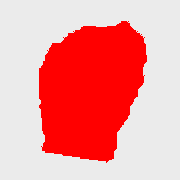}
    \includegraphics[width=.08\textwidth]{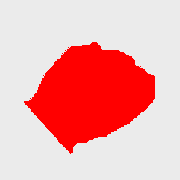}
    \includegraphics[width=.08\textwidth]{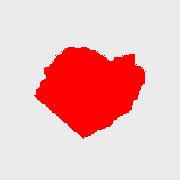}
    \includegraphics[width=.08\textwidth]{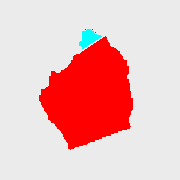}\\
    \vspace{0.08cm}

    \includegraphics[width=.08\textwidth]{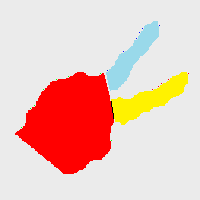}
    \includegraphics[width=.08\textwidth]{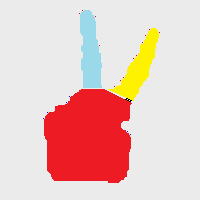}
    \includegraphics[width=.08\textwidth]{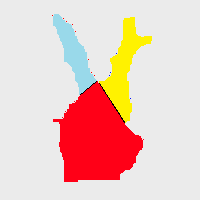}
    \includegraphics[width=.08\textwidth]{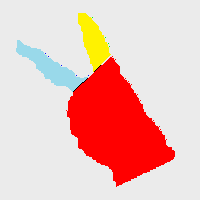}
    \includegraphics[width=.08\textwidth]{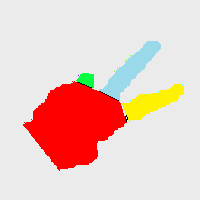}\\
	\vspace{0.1cm}

    \includegraphics[width=.08\textwidth]{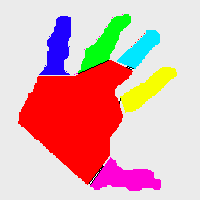}
    \includegraphics[width=.08\textwidth]{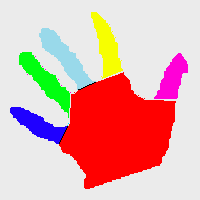}
    \includegraphics[width=.08\textwidth]{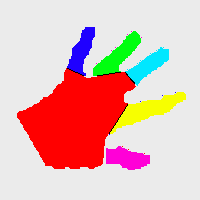}
    \includegraphics[width=.08\textwidth]{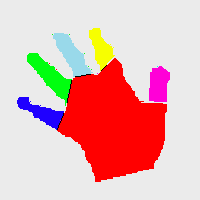}
    \includegraphics[width=.08\textwidth]{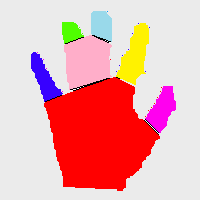}
    \end{center}
    \vspace{-.2cm}
       \caption{
       Hand gesture decomposition results. Although shapes in the last column are not decomposed perfectly, they can still be discriminated from other categories by counting the number of parts.
       }
    \label{fig:handSeg}
    \end{figure}

    We now demonstrate the efficiency of the proposed algorithm by applying it to hand gesture recognition.
    Following
    the setting in \cite{2011ICCV_NearConvex},
    the task is to recognize hand gestures of three categories, namely Rock, Paper and Scissors.
    The data set, including both color images and depth maps, is collected by a Kinect depth camera.
   Each category has 100 samples collected from 10 subjects. As shown in Fig. \ref{fig:handGesture}, with the help of depth maps, hand shapes are easy to be segmented, although not perfectly, from the scenes. We then decompose them by the proposed algorithm.
    Fig.\ \ref{fig:handSeg} shows some decomposition results of hand gestures from different subjects under various scale, orientation and illumination conditions.
    Similar to \cite{2011ICCV_NearConvex}, we classify a gesture to Rock if $k\leq2$, Paper if $k\geq5$, and Scissors otherwise, where $k$ is the number of parts. With the setting of $t_{\text{DCE}} = 1.2$, the proposed method perfectly distinguishes all of these 300 gestures into correct categories comparing with \cite{2011ICCV_NearConvex} whose mean recognition accuracy is 94.7\% on a even smaller set.

\section{Conclusion}
\label{sec:con}

    In this work, we have proposed a new flexible shape decomposition approach based on
    the short-cut rule, which roots in psychology.
    In other words, we have proffered a computational procedure of the short-cut
    rule, and applied it to 2D shape decomposition.
    An important component of the proposed approach is that
    we divide the potential cuts of a shape into two types
    according to the number of \mm points that they have.
    We then examine them in order from shortest to longest to determine part-cuts, based on the short-cut rule.

    Experiments show that our
    method can separate shapes into intuitive parts with low time complexity.
    Our approach can also be improved easily by introducing more constraints
    from visual observations.
    Although a few parameters are involved in our model, we empirically show that
    the final performance is not very sensitive to these parameters in a large range.
    We will explore the possibility of automatically tuning these parameters
    by computing a measure of salience for
    part-cut hypotheses with the definition proposed by  \cite{Hoffman2}.

\appendices

\section{Discrete Curve Evolution}
\label{sec:append_DCE}

The Discrete Curve Evolution (DCE) algorithm was introduced in \cite{DCE7} to obtain shape hierarchy for multiscale shape analysis.
Contours of shapes are usually distorted by digitization noise.
DCE regards a shape contour as a polygon with a large number of vertices,
and simplifies it with an evolution process to eliminate distortions.
The evolution process is done according to a criterion $K(\cdot,\cdot)$ measuring the significance of two consecutive edges' contribution to the shape:
\begin{equation}
K(s_1,s_2)=\frac{\beta(s_1,s_2)l(s_1)l(s_2)}{l(s_1)+l(s_2)},
\end{equation}
where $\beta(s_1,s_2)$ is the turn angle at the common vertex of $s_1,s_2$, $l(\cdot)$ is the normalized length of an edge.
The higher value $K(s_1,s_2)$ is, the larger $s_1, s_2$ contribute to the shape.
In each evolutional step, the pair of consecutive edges with minimum $K$ value is selected to be replaces with a new line segment joining the endpoints of them.
The evolution will converge to a convex polygon.
If a termination threshold $t_\text{DCE}$ is given, it stops when no $K$ value is less than $t_\text{DCE}$.
An illustration of the procedure is shown in Fig.~\ref{fig:deer}, where the simplified polygon (in blue dashed lines) of the deer getting rougher as $t_\text{DCE}$ inceases.
In general, DCE is a greedy approach to simplify the contour while keeping its geometric structure as much as possible.
See \cite{DCE7, DCE17} for more details about the  DCE algorithm.

\section{Illustration of Observations 3}
\label{sec:append_obs3}

    \begin{figure}[t]
    \begin{center}
    \includegraphics[width=0.35\textwidth]{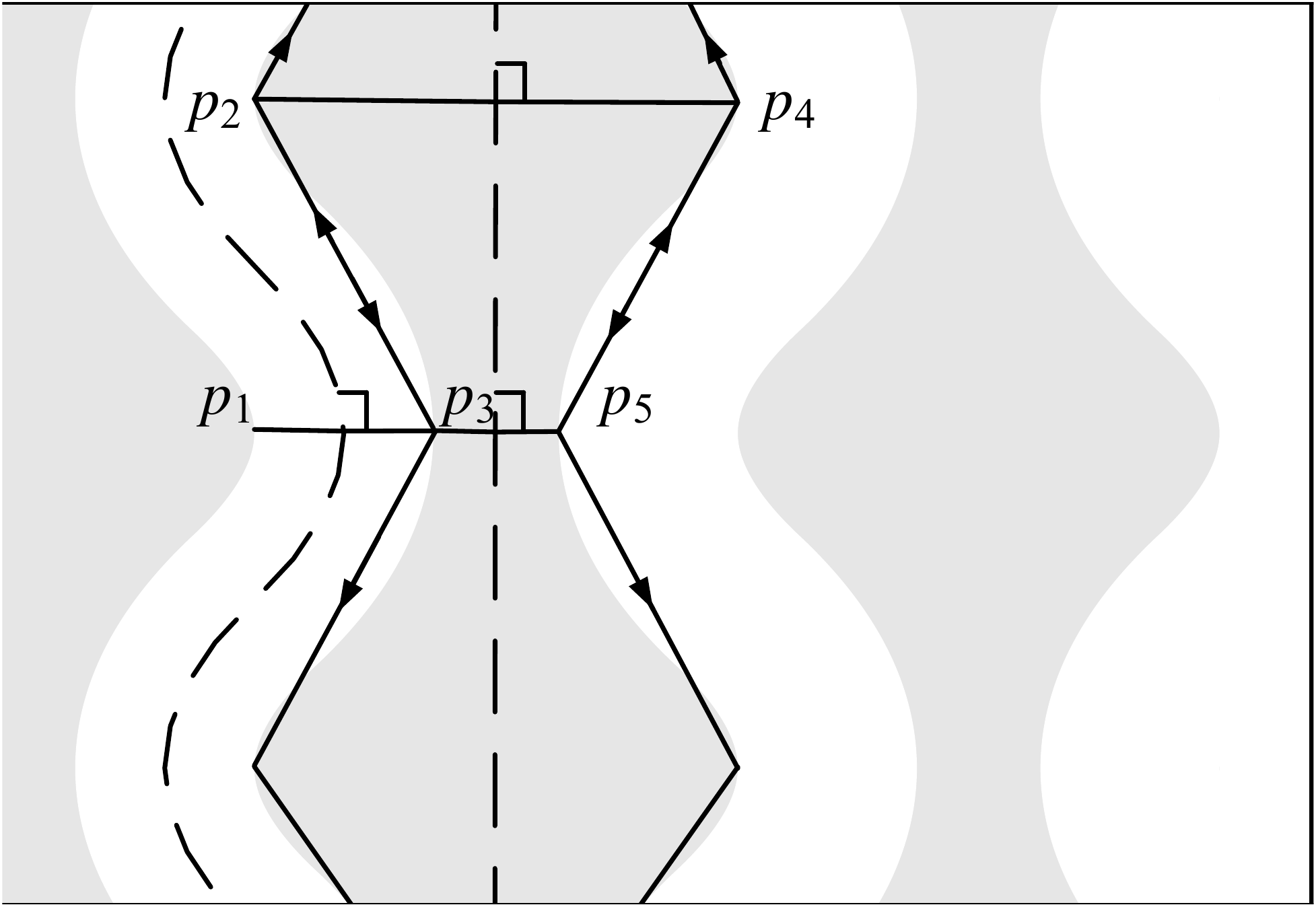} %
    \end{center}
    \vspace{-.2cm}
       \caption{Parallelism (white) and co-circularity (grey).
       The grey
      ``dumbbell" is simplified into a polygon by DCE. $p_1p_3,p_2p_4$ and $p_3p_5$
       both
       are orthogonal to the local symmetry axis (dashed line),
       while $p_3p_5$ is more likely to be a part-cut. }
    \label{fig:symmety}
    \end{figure}

    We  here give a detailed explanation of the term of ``expanding" in Observation 3.
    In an arbitrary shape, two types of local symmetry---parallelism and co-circularity---may appear.
    As illustrated in Fig. \ref{fig:symmety}, the outline of the white ``wiggle" is parallel,
    and the gray ``dumbbell" is mirror symmetric, or co-circular.
    The length of the rib along the axis of local symmetry,
    or the \emph{local width}, of the wiggle keeps unchanged,
    while it varies with the co-circular outline of the dumbbell.
    Previous research \cite{shock14,rom1993hierarchical}
    suggested that the human vision system does not tend to parse the wiggle into parts.
    Thus only the symmetry of co-circularity is considered here.
    Giblin and Brassett \cite{1985symmetry} proved that the axes of local symmetry generally are smooth curves.
    It means that the local width of the (co-circular) boundary, starting from a cut which is orthogonal to the axis of local symmetry, may vary continuously in two trends---expanding or shrinking.
    As shown in Fig. \ref{fig:symmety}, departing from $p_3 p_5$, either upwards or downwards, the trend is expanding, while it is shrinking from $p_2 p_4$.
    In general, ``expanding" and ``shrinking" are the description of the varying trends of local width near the part-cut hypothesis.

\section*{Acknowledgments}

This work was in part funded by grants NSFC 61033008,  61103080, 61272145 and 61402504, ARC FT120100969.

{\footnotesize
\bibliographystyle{ieee}
\bibliography{shape}
}

\end{document}